%% file: paper.tex
\documentclass[]{amap}

\usepackage[toc,page,header]{appendix}
\usepackage{minitoc}
\usepackage{solarized-light}
\usepackage{booktabs}
\usepackage{multirow}
\usepackage{graphicx}
\usepackage{array}
\usepackage{siunitx,array}
\usepackage[table]{xcolor}
\usepackage{makecell,array,booktabs}
\usepackage{makecell}
\usepackage{framed}
\usepackage{amssymb}
\usepackage{bbding}
\usepackage{adjustbox}
\usepackage{amsmath}
\usepackage{placeins}
\usepackage[colorinlistoftodos]{todonotes}
\usepackage{longtable}
\usepackage{hhline}
\usepackage{fancyvrb}
\usepackage{float}
\usepackage{fvextra}
\usepackage{CJKutf8}
\usepackage{multicol}
\usepackage{tablefootnote}
\usepackage{threeparttable}
\usepackage{tabularx}
\usepackage{mdframed}
\usepackage{subcaption}
\usepackage[usestackEOL]{stackengine}
\usepackage{enumitem}
\usepackage{microtype}
\usepackage{url}
\usepackage{amsthm}

\newcommand{\commentout}[1]{}
\renewcommand{\paragraph}[1]{\noindent\textbf{#1}\hspace*{1em}}
\setlist[itemize]{leftmargin=15pt}


\providecommand{\fitwidth}[1]{%
  \resizebox{\ifdim\width>\linewidth \linewidth\else \width\fi}{!}{#1}}
  
\RequirePackage{xspace}
\makeatletter
\DeclareRobustCommand\onedot{\futurelet\@let@token\@onedot}
\def\@onedot{\ifx\@let@token.\else.\null\fi\xspace}

\makeatother

\graphicspath{{figs/}}

\title{ABot-ER: Unified Embodied Reasoning through Agentic Data–Model Co-Evolution}

\vspace{-10pt}

\abstract{

We introduce \textbf{ABot Embodied Reasoning (ABot-ER)}, a unified embodied reasoning foundation model that breaks the tension between breadth and depth in embodied reasoning. Embodied intelligence demands the joint mastery of perception, planning, and execution, yet reconciling generality with specialization across this spectrum remains an open challenge. Within a single unified model, ABot-ER commands specialist-level depth across all three competencies, spanning spatial, scene, and temporal reasoning in perception; long-horizon action organization in planning; and point grounding and affordance in execution. 
The \textbf{Agentic Data-Model Self-Evolution Engine} continuously adapts data strategies to the model's evolving capability state. The \textbf{Task-Conditioned Adaptive Training Strategy} uses task tokens to activate lightweight low-rank residuals over a shared backbone, separating updates from heterogeneous output formats while retaining shared embodied knowledge. 
Across a systematic evaluation on twelve benchmarks covering three core competencies and five capability dimensions, ABot-ER achieves \textbf{state-of-the-art overall performance} across perception, planning, and execution within a unified parametric framework, demonstrating that breadth and depth can coexist in unified embodied reasoning.

\textbf{Project Page:} \url{https://github.com/amap-cvlab/ABot-ER}

}

\begin{document}
\maketitle
\vspace{-4pt}




{%
\vspace{-14pt}

\centering
\includegraphics[width=\linewidth,height=0.5\textheight,keepaspectratio]{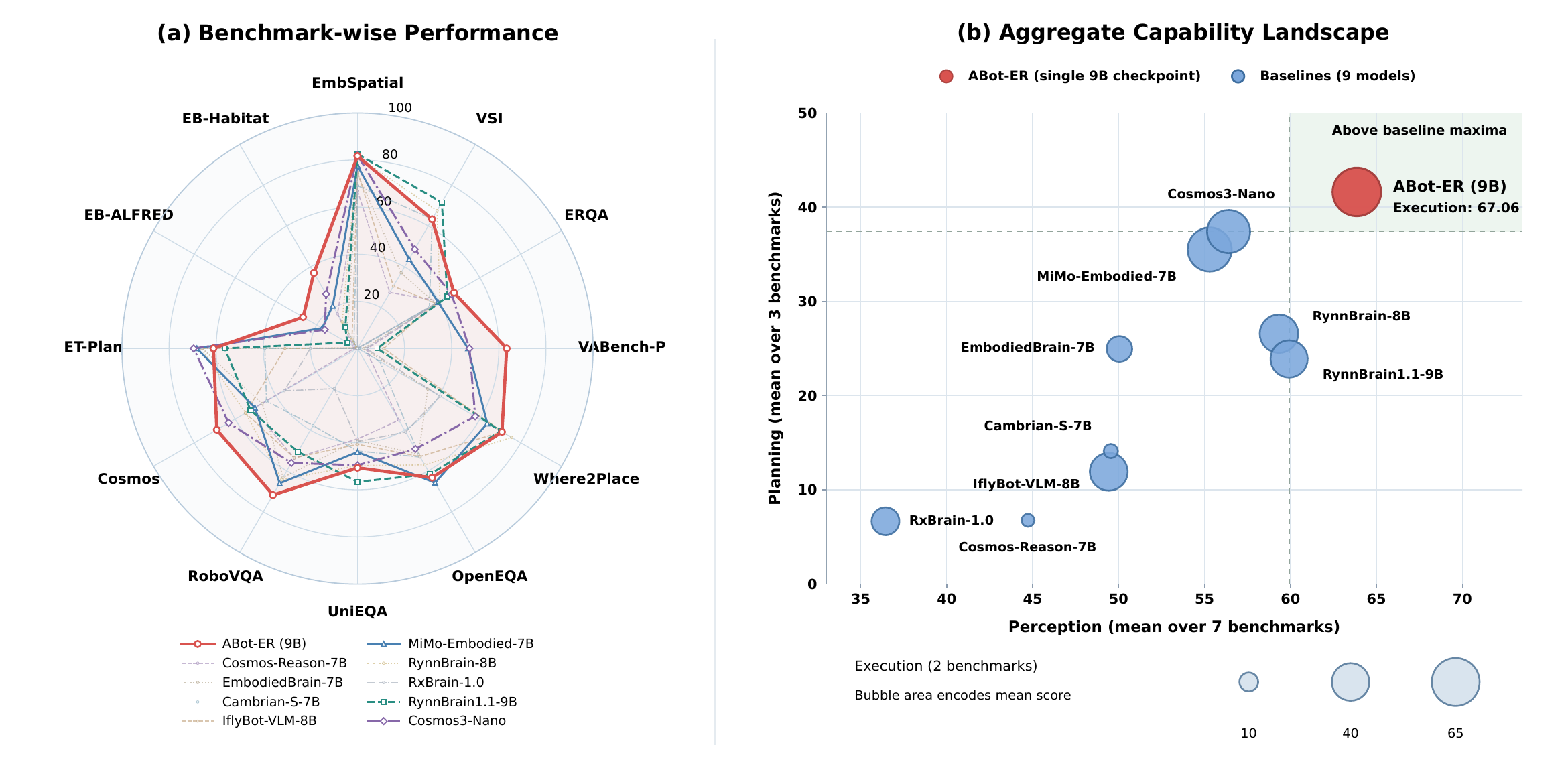}

\captionsetup{type=figure}
\small
\captionof{figure}{\textbf{Embodied reasoning performance of ABot-ER.} Left: the twelve benchmarks against nine public baselines. Right: the perception and planning means as axes, with marker area encoding the execution mean. From one 9B checkpoint, ABot-ER is the only model above the best baseline value on both axes.}
\label{fig:teaser}
\par}

\newpage
\tableofcontents
\newpage

\input{sections/introduction}
\input{sections/embodied_reasoning}
\input{sections/related}

\input{sections/architecture}  

\input{sections/data_agent_v3}

\input{sections/er-experiment}

\input{sections/conclusion}

\clearpage

\input{sections/contribution}
\clearpage

\clearpage

\bibliographystyle{plainnat}
\bibliography{main}

\begin{appendices}
\input{sections/appendix_engine}

\end{appendices}

\end{document}

%% file: sections/introduction.tex
\section{Introduction}
\label{sec:intro}

Embodied intelligence studies agents that act on instructions in the physical world. An instruction such as \emph{rinse off a ladle and move it to the table} requires more than recognizing objects: determining which objects and scene states are relevant, what sequence of actions is needed, and where each action should be directed. These interdependent judgements span perception, planning, and execution, and together constitute \emph{embodied reasoning}. Execution here means identifying grounded interaction targets for a downstream action model rather than generating motion commands. Moving beyond a fixed task requires coordinating these judgements as tasks, scenes, and embodiments change, which makes embodied reasoning central to general-purpose behaviour in the physical world.

Embodied reasoning (ER) models serve as a reasoning interface between multimodal observations and downstream control. Recent embodied foundation models reflect a broader shift from isolated perception or control tasks toward reasoning across multiple stages of physical interaction~\cite{gemini2025robotics,robobrain2025report,nvidia2025cosmosreason,hao2025mimo,rynn2026brain}. An ER model infers spatial relations, scene states, and temporal changes, translates high-level goals into ordered action plans, and identifies where interaction should occur~\cite{majumdar2024openeqa,du2024embspatial,yang2024vsibench,ahn2022saycan,yuan2024robopoint}. Its outputs are task-relevant assessments, action plans, and interaction targets that downstream systems use to guide and parameterize execution.

However, existing ER models still exhibit uneven capabilities across perception, planning, and execution. Unifying these competencies raises two related challenges, one about data and one about optimization: which types of supervision should be prioritized at each training stage, and how heterogeneous task outputs can be learned jointly through shared parameters while limiting interference. The first, which we term \textbf{Data-Model Drift}, arises because capabilities develop at different rates: the relative utility of each type of training data can change over the course of training, so a fixed mixture may become less well aligned with the model's evolving learning needs~\cite{xie2023doremi}. This mismatch can arise even when the underlying corpus remains unchanged. The second, which we term \textbf{Modality Entanglement}, arises when heterogeneous outputs, namely semantic descriptions, structured action sequences, and spatial coordinates, impose competing demands on shared parameters, so an update that benefits one output type may degrade another~\cite{yu2020pcgrad}. Section~\ref{sec:architecture} fixes the boundary these outputs are separated along.


ABot-ER addresses data adaptation and optimization interference with an \textbf{Agentic Data-Model Self-Evolution Engine} and a \textbf{Task-Conditioned Adaptive Training Strategy}, respectively. The engine pursues recursive self-improvement (RSI) at the level of training data, and turns corpus construction into a closed loop driven by evaluation feedback, with the aim of improving performance across capability dimensions: the current model's errors are diagnosed and attributed to specific capability gaps, targeted supervision is constructed to address them and audited for relevance and quality, the training-data mixture across capability dimensions is adjusted, the model is trained on the revised corpus, and the resulting evaluation opens the next round. What recurses is the corpus, not the learning algorithm or the architecture, and what enters each round is decided by deterministic gates rather than by the model itself. The loop is governed by continuous screening of each corpus revision for overlap with evaluation data, under quality-control rules calibrated through expert review in the initial rounds; subsequent rounds proceed automatically under those rules. The training strategy uses a task token to activate lightweight low-rank residuals selected by the task type of each training record. All tasks share a backbone for cross-task representation learning, while the residuals provide dedicated capacity for type-specific adaptation, reducing interference between heterogeneous outputs. We systematically evaluate ABot-ER on twelve benchmarks covering three core competencies and five capability dimensions. Within a unified parametric framework, it achieves \textbf{state-of-the-art overall performance}, with the highest aggregate score in each of perception, planning, and execution, while remaining competitive on individual benchmarks, demonstrating that breadth and depth can coexist in unified embodied reasoning.

The main contributions of this report are as follows:
\begin{itemize}
    \item We design the \textbf{Agentic Data-Model Self-Evolution Engine}, which realizes RSI over the training corpus by coupling evaluation-driven capability diagnosis with targeted data revision and iterative training. By adapting supervision and training-data mixtures to the model's evolving capability needs, the engine mitigates \textbf{Data-Model Drift}. The recursion is bounded rather than open-ended, and Section~\ref{sec:experiments} reports where it converges.
    \item We propose the \textbf{Task-Conditioned Adaptive Training Strategy}, which uses task tokens to activate dedicated lightweight low-rank residuals for semantic and action outputs while retaining a shared backbone for cross-task learning. This provides output-specific adaptation capacity to mitigate the optimization interference associated with \textbf{Modality Entanglement}.
    \item We systematically evaluate ABot-ER on twelve benchmarks spanning three core competencies and five capability dimensions. Within a unified parametric framework, it achieves \textbf{state-of-the-art overall performance}, with the highest aggregate score in each of perception, planning, and execution.
    \item We demonstrate the \textbf{reusability of the engine-produced corpus}: from the same ancestor checkpoint, a single post-training run on this corpus matches the six-round iterative model in mean performance across the twelve benchmarks, with a fifth of the accumulated post-training. The same corpus and post-training recipe are also applied to backbones ranging from 2B to 27B parameters, without rerunning the data-revision loop for each backbone. To support further research, we will open-source our \textbf{pre-training and post-training datasets}, together with a \textbf{multi-scale ABot-ER model family} comprising 2B, 4B, 9B, and 27B models, as well as a 35B-A3B MoE model.
\end{itemize}

The remainder of this report is organized as follows: Section~\ref{sec:er_role} states the role of an embodied reasoning model in a robot system and the decisions ABot-ER is designed to make; Section~\ref{sec:related} reviews related work; Section~\ref{sec:architecture} formalizes the multi-competency unification problem and presents the unified model architecture, with the Task-Conditioned Adaptive Training Strategy in Section~\ref{sec:tcat}; Section~\ref{sec:engine} details the Agentic Data-Model Self-Evolution Engine; Section~\ref{sec:experiments} reports experimental results and in-depth analysis; Section~\ref{sec:conclusion} concludes and discusses future directions.

%% file: sections/embodied_reasoning.tex
\section{The Role of an Embodied Reasoning Model}
\label{sec:er_role}

\begin{figure*}[t]
  \centering
  \includegraphics[width=\textwidth]{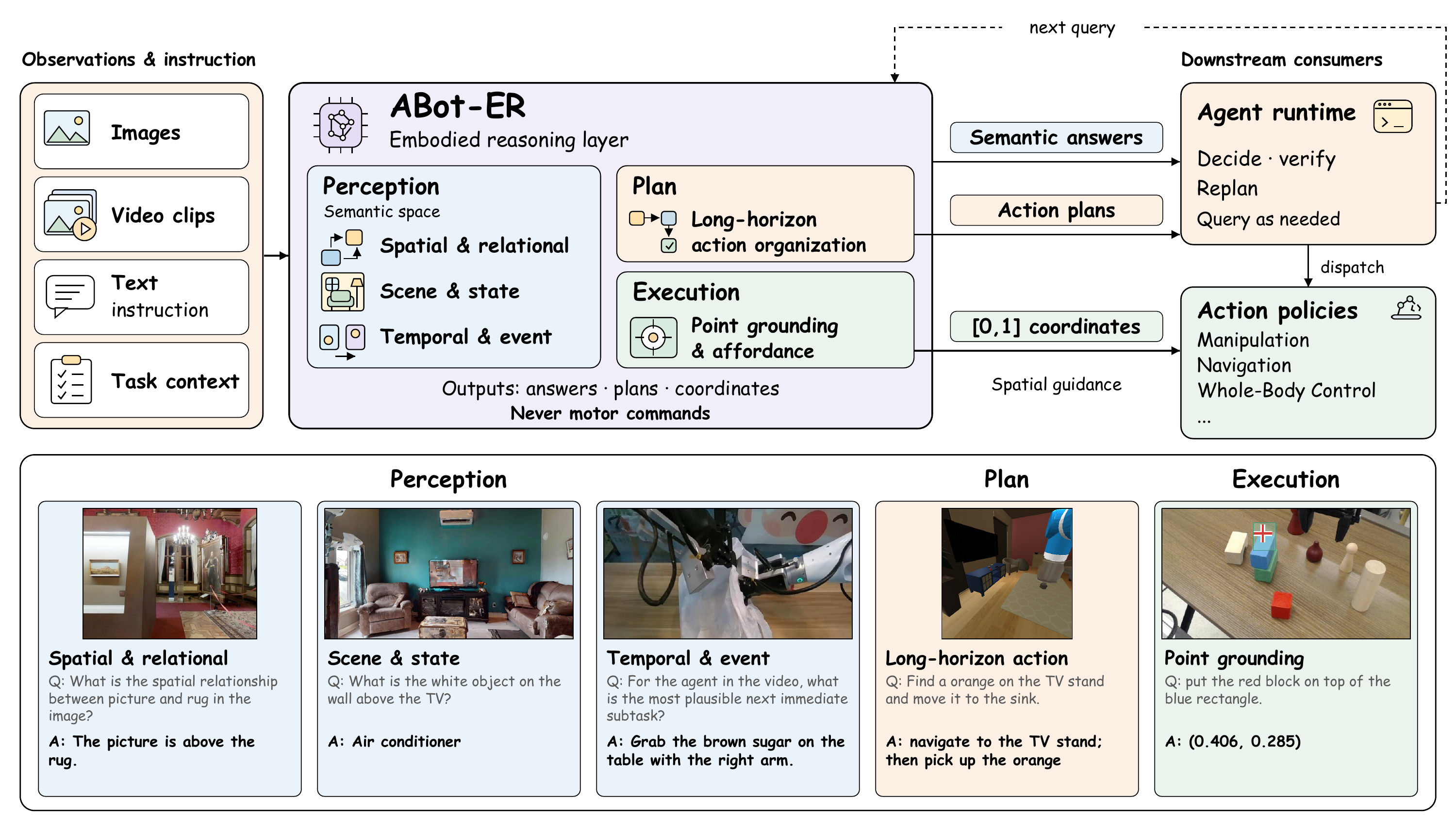}
  \caption{The embodied reasoning layer between what a robot sees and what it does. It takes observations and an instruction and returns three kinds of answer: semantic assessments, ordered action plans, and normalized image coordinates. The agent runtime uses the assessments and plans to manage task execution, while the coordinates specify spatial targets for the downstream policies; the model itself emits no motor commands. Bottom: one official test case per capability dimension, drawn from the evaluation splits of Section~\ref{sec:experiments} and grouped by the competency it belongs to. Each card pairs the benchmark question with ABot-ER's own answer. All five answers were scored correct: the planning case succeeded over the full episode, and the predicted point fell inside the official target mask, which is shaded in the image.}
  \label{fig:er_role}
\end{figure*}

An embodied reasoning model supplies task-relevant judgements throughout a physical interaction. Given observations and a task instruction, it reasons about spatial relations, scene states, temporal events, action sequences, and interaction targets, which together support both deciding what to do next and assessing what has already happened. Depending on the query, it returns a semantic assessment, an ordered action plan, or grounded image coordinates, and it does not generate motor commands. The agent runtime manages task progression: it decides when to query the model, acts on the returned assessments and targets, and schedules and supervises the proposed plans. Downstream action policies carry out the motion, navigation~\cite{abotn0_2025,abotn1_2026} and manipulation~\cite{yang2026abotm0vlafoundationmodel,chen2026abotm05unifiedmobilityandmanipulationworld} alike. ABot-ER instantiates this role (Figure~\ref{fig:er_role}; physical deployment in Section~\ref{sec:realrobot}). This section defines the five capability dimensions the role requires and explains why we serve them from one model.

\subsection{What the Model Answers for the Robot}
\label{sec:er_questions}

We organize these reasoning requirements into five capability dimensions, summarized in Table~\ref{tab:er_questions_en} and formalized in Section~\ref{sec:architecture}. The first three, spatial and relational reasoning, scene and state understanding, and temporal and event reasoning, fall under perception. The remaining two address planning and execution-oriented grounding, and we refer to their outputs as action-space outputs, distinguished from semantic outputs by their task function and interface requirements.

The five form one set rather than five separate problems, because a single action draws on several of them at once. Placing an object needs free space (spatial), a clear surface (state), and a release point (point grounding); recovering from a failed grasp needs the failure recognized (temporal), the object re-localized (spatial), and the plan revised (long-horizon). The answers must also agree with one another: a release point is only useful if it lies in the region the state assessment just called clear.

\begin{table*}[!htbp]
\centering
\caption{The five capability dimensions and the form of each answer. Examples follow one task, the Block Arithmetic demonstration of Section~\ref{sec:realrobot}: a participant seated across the table lays out $6+9$ with blocks whose value is given by the number of dots on them, oriented toward the participant, and the robot places two blocks that read as the result. Directional descriptions are stated from the participant's viewpoint, whereas coordinates follow the normalized robot-camera image protocol of Section~\ref{sec:architecture}.}
\label{tab:er_questions_en}
\small
\renewcommand{\arraystretch}{1.18}
\setlength{\tabcolsep}{5pt}
\begin{tabular}{@{}p{0.21\textwidth}p{0.19\textwidth}p{0.38\textwidth}p{0.15\textwidth}@{}}
\toprule
\textbf{The robot needs to know} & \textbf{Capability} & \textbf{Example} & \textbf{Answer form} \\
\midrule
Where things are and how they are arranged & Spatial and relational reasoning & Which block is the left operand from the participant's viewpoint, given that the participant sits across the table and the expression therefore reads right to left in the robot's camera image & Choice, number, or short text \\
\addlinespace[2pt]
What state the scene is in & Scene and state understanding & How many dots each block carries and which operator symbol is shown: $6+9$ & Short text or choice \\
\addlinespace[2pt]
What just happened, and whether it worked & Temporal and event reasoning & From the frames of the arm's placement motion, whether the five-dot block was placed on the table or dropped during transfer & Choice: placed, or dropped in transfer \\
\addlinespace[2pt]
What to do, in what order, under which preconditions & Long-horizon action organization & Place the one-dot block that stands for the tens digit, then, once it is down, place the five-dot ones digit to its right, so that the pair reads $15$ & Ordered plan in the executor's schema \\
\addlinespace[2pt]
Exactly where to grasp, place, or act & Point grounding and affordance & A grasp point on the five-dot block, and a placement point inside a free region to the right of the tens digit large enough to accommodate it & Normalized image coordinates in $[0,1]$ \\
\bottomrule
\end{tabular}
\end{table*}

\subsection{Why One Dedicated Model}
\label{sec:er_contribution}

Prior work has demonstrated interaction-point prediction~\cite{yuan2024robopoint} and spatial reasoning~\cite{chen2024spatialvlm} in vision-language models, as well as reasoning-augmented action generation in vision-language-action models~\cite{brohan2023rt2,gemini2025robotics}. The design question is therefore not whether an individual question of this kind can be answered, but how the five dimensions are brought together behind one interface that an agent runtime and downstream policies can build on. We address this question with a model trained jointly across the five dimensions under explicit output protocols, specified in Sections~\ref{sec:architecture} and~\ref{sec:protocol}. Two commitments follow from the role: the model must report on outcomes as well as propose actions, and it must serve all five dimensions from one checkpoint.

Acting on an answer is only half of the loop; the other half is knowing whether the action worked. By assessing scene states and temporal events, such as whether an object has been placed or which event occurred first, the model provides the information from which an agent runtime decides whether to proceed, retry, or re-plan. On the physical platform of Section~\ref{sec:realrobot}, the model reads the tabletop before the arm acts, recovering the participant's expression, and again afterwards, identifying the digit sequence represented by the placed blocks.

The second commitment concerns where the answers come from. ABot-ER serves the five dimensions from a single 9B-parameter checkpoint. The shared backbone provides a common representational basis for spatial assessment, planning, and target grounding, while task-conditioned low-rank residuals provide capacity for task-family-specific adaptation (Section~\ref{sec:architecture}). At deployment this means one model to serve rather than a specialist per dimension, and the physical system of Section~\ref{sec:realrobot} runs it on a single NVIDIA RTX 5090D.

%% file: sections/related.tex
\section{Related Work}
\label{sec:related}

We organize prior work along four axes most relevant to ABot-ER: (i) embodied vision-language foundation models that consolidate multiple capabilities into unified embodied models, (ii) embodied perception and interaction grounding, (iii) task planning and feedback-guided execution, and (iv) adaptive data construction and multi-task training, which is the background for the report's two methodological components. Comprehensive treatments of the broader embodied AI landscape can be found in recent surveys~\cite{xu2024robotics_survey,liu2024embodied_survey}.

\subsection{Embodied Vision-Language Foundation Models}

Earlier embodied learning studies often focused on specific task settings, including embodied question answering~\cite{das2018eqa}, manipulation~\cite{shridhar2021cliport}, and vision-and-language navigation~\cite{anderson2018vln}. Their different task interfaces and supervision objectives motivate input-output alignment when constructing a unified model. A recent line of work consolidates multiple embodied capabilities under a single vision-language model. Two useful distinctions are how sensor observations enter the model and whether it directly generates robot-control actions. PaLM-E~\cite{driess2023palme} embeds sensor observations directly into the token space, so that a large language model performs cross-task embodied reasoning over them. RT-2~\cite{brohan2023rt2} places robotic action in the same output vocabulary as vision-language pretraining within a single autoregressive framework. EmbodiedGPT~\cite{mu2023embodiedgpt} uses embodied chain-of-thought for high-level planning and extracts planning-related features to support closed-loop control, keeping plan generation and control in separate stages. LEO~\cite{huang2023leo} integrates 3D vision, language, and action modeling in a unified embodied agent, and the RoboBrain~\cite{robobrain2025report} series builds multimodal foundation models covering spatial understanding, temporal reasoning, and task planning through progressive multi-stage training. ABot-ER follows this line of multi-capability modeling, but exposes semantic assessments, action plans, and interaction targets rather than directly generating motor commands.

\subsection{Embodied Perception and Interaction Grounding}

Embodied perception extracts structured information from visual observations in support of physical interaction decisions. Interaction grounding builds on this visual understanding but is treated as execution-oriented reasoning in our capability taxonomy (Section~\ref{sec:er_questions}). In spatial relation reasoning, SpatialVLM~\cite{chen2024spatialvlm} endows vision-language models with quantitative spatial reasoning through large-scale synthetic spatial question answering, and SpatialRGPT~\cite{cheng2024spatialrgpt} combines region-level reference with depth information to improve relative position judgement. On the evaluation side, EmbSpatial-Bench~\cite{du2024embspatial} assesses spatial relations from an embodied viewpoint, whereas VSI-Bench~\cite{yang2024vsibench} assesses spatial understanding from video. In interaction grounding, VRB~\cite{bahl2023vrb} learns affordance representations from human videos, ManipLLM~\cite{li2023manipllm} predicts manipulation-oriented end-effector poses from RGB observations and text prompts, and Where2Place, introduced with RoboPoint~\cite{yuan2024robopoint}, evaluates valid placement prediction under spatial constraints. In scene state understanding, OpenEQA~\cite{majumdar2024openeqa} is an open-ended embodied question-answering benchmark, and RoboVQA~\cite{sermanet2023robovqa} constructs large-scale question answering over robot and human videos, covering planning, success judgement, affordance, past description, and future prediction. In temporal reasoning, Cosmos-Reason1~\cite{nvidia2025cosmosreason} targets physical common sense and embodied reasoning over video. These studies emphasize different capabilities and use task-specific supervision and output conventions, so integrating them requires explicit alignment of task definitions and interfaces.

\subsection{Task Planning and Feedback-Guided Execution}

Task planning translates high-level goals into action choices and sequences while accounting for feasibility, ordering constraints, and feedback. The paradigm of LLM-as-planner established several routes to this: SayCan~\cite{ahn2022saycan} combines semantic knowledge with value-function feasibility for task decomposition; ProgPrompt~\cite{singh2022progprompt} and Code-as-Policies~\cite{liang2022codeaspolicies} cast planning as code generation, using program structure for sequencing and control flow; and Inner Monologue~\cite{huang2022innermonologue} introduces environment feedback as an online correction signal, so that plans are revised during execution rather than fixed in advance. At the execution level, benchmarks such as ALFRED~\cite{shridhar2020alfred} and simulation platforms such as Habitat~\cite{savva2019habitat} require a model to select actions conditioned on successive observations during interaction. These approaches use different perception interfaces, plan representations, and feedback mechanisms, providing complementary foundations for unified embodied modeling.

\subsection{Adaptive Data Construction and Multi-Task Training}

The two methodological components of this report, one concerning which supervision to train on and one concerning how heterogeneous outputs share parameters, each have a line of prior work.

\paragraph{Data construction, self-improvement, and mixture weighting.} Self-Instruct~\cite{wang2022selfinstruct} generates and filters instruction data with the model itself. Self-Rewarding Language Models~\cite{yuan2024selfrewarding} use model-generated judgements as the reward signal in iterative training. RoboCat~\cite{bousmalis2023robocat} self-improves from self-generated robotic interaction data, so that data and model advance together across generations. DoReMi~\cite{xie2023doremi} optimizes data-domain weights with a proxy model and applies the resulting mixture to subsequent training. ABot-ER's engine differs from these in what drives the revision and in what is revised. The engine diagnoses weaknesses at the capability-dimension level from evaluation evidence and revises both supervision content and cross-dimension mixtures within each round, with capability gains evaluated alongside retention. Whereas DoReMi learns sampling weights over existing data domains, our engine also constructs and revises supervision in response to capability-level diagnosis.

\paragraph{Parameter-efficient adaptation and multi-task interference.} Adapter modules~\cite{houlsby2019adapter} add small task-specific components to a pretrained network, while LoRA~\cite{hu2021lora} represents weight updates through low-rank factors. In their standard transfer-learning formulations, the pretrained backbone remains frozen. A separate line addresses multi-task interference through gradient-based optimization: PCGrad~\cite{yu2020pcgrad} projects away conflicting components of task gradients, while CAGrad~\cite{liu2021cagrad} maximizes the worst-case local improvement across tasks within a neighborhood of the mean task gradient. Our Task-Conditioned Adaptive Training Strategy uses low-rank residual parameterization but jointly trains the shared backbone and residual branches, without explicit gradient-conflict correction. A task token selects the branch according to the sequence's output regime (semantic or action-space) rather than allocating a separate branch to each benchmark or capability dimension (Section~\ref{sec:architecture}).

\vspace{0.5em}

Against this background, ABot-ER studies unified embodied reasoning across perception, planning, and execution-oriented grounding. Its data engine couples capability diagnosis with targeted supervision revision and capability-aware mixture adjustment over multiple training rounds, with gains evaluated alongside retention. Its task-conditioned training strategy provides dedicated low-rank residual capacity for semantic and action-space outputs while retaining a shared backbone. These components are studied together as a framework for adaptive data construction and joint training across the five capability dimensions.

%% file: sections/architecture.tex
\section{ABot-ER}
\label{sec:architecture}


\begin{figure*}[ht]
  \centering
  \includegraphics[width=\textwidth]{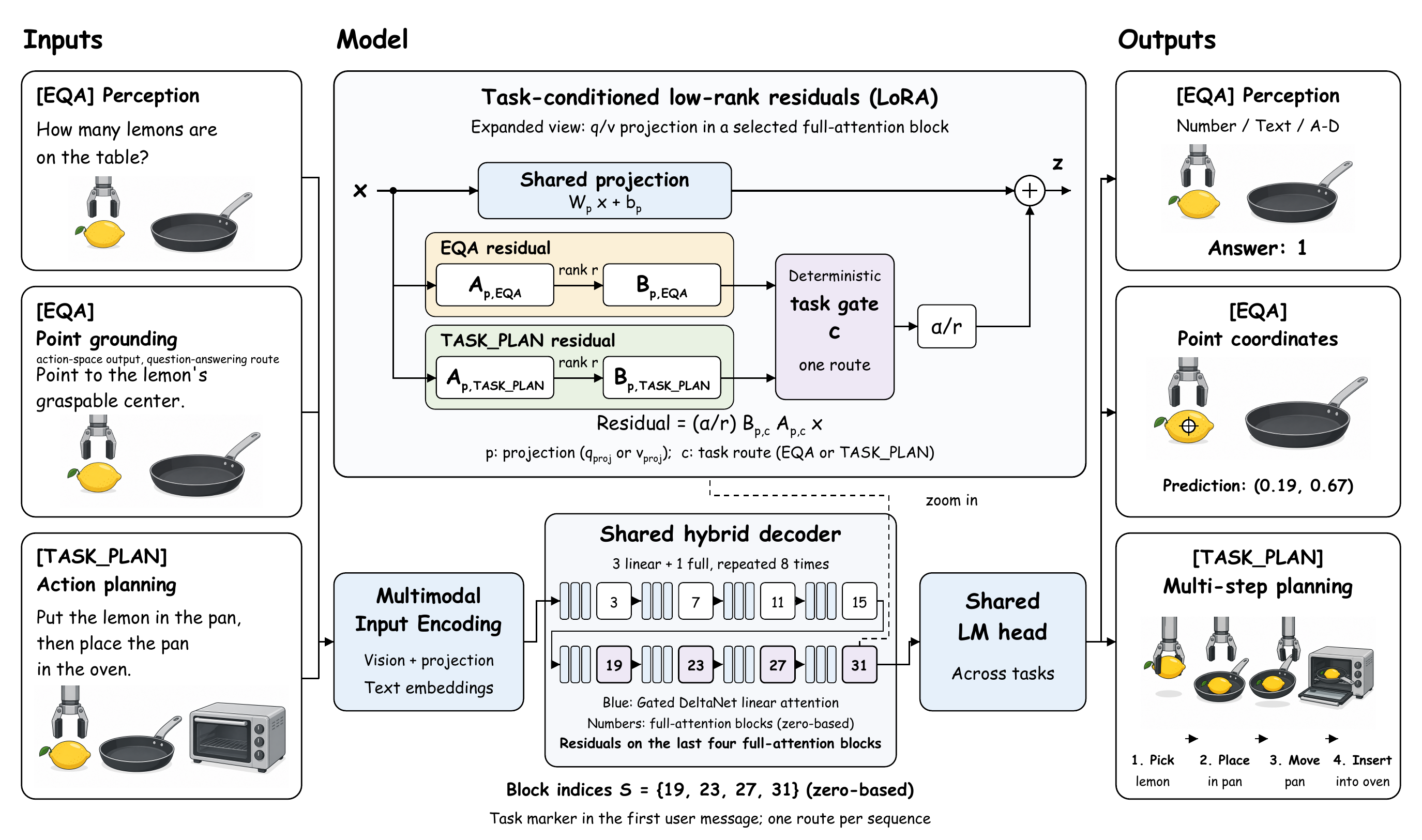}
  \caption{Model architecture.}
  \label{fig:dhta}
\end{figure*}

\subsection{Capability Coverage and Unified Modeling Objective}

We refine the three core capabilities of perception, planning, and execution into five capability dimensions:
\begin{itemize}
    \item \textbf{Spatial and Relational Reasoning (Perception):} Judging direction, distance, and multi-view spatial layouts among objects.
    \item \textbf{Scene and State Understanding (Perception):} Recognizing object attributes, inferring interaction states, and performing open-ended scene question answering.
    \item \textbf{Temporal and Event Reasoning (Perception):} Recognizing actions and events in video, determining temporal order, and detecting state changes.
    \item \textbf{Long-Horizon Action Organization (Planning):} Generating ordered action plans under task preconditions and cross-step state constraints given high-level goals, and selecting subsequent actions from available observations and interaction history.
    \item \textbf{Point Grounding and Affordance (Execution):} Localizing manipulation targets and predicting placement positions under affordance constraints, expressed as image coordinates for downstream action policies.
\end{itemize}
The evaluation benchmarks and metrics corresponding to each capability dimension are listed in Table~\ref{tab:metrics_en}.

Although the five capability dimensions differ in their reasoning requirements and output formats, they share core embodied knowledge at the foundational level, including object, spatial, state, and affordance representations. We cover all five dimensions within a single unified model and formulate the supported tasks within a shared conditional generation framework. Let $\mathbf{x}$ denote the available task context, which may consist of text alone or include visual observations. State descriptions, action histories, and environmental feedback are incorporated when provided by the task. Conditioned on $\mathbf{x}$ and the task token $c$, whose values and assignment rule are given below, ABot-ER generates the required response $\mathbf{y}$ autoregressively:
\begin{equation}
    p_\theta(\mathbf{y}\mid\mathbf{x},c)
    =
    \prod_{t=1}^{T}
    p_\theta(y_t\mid\mathbf{x},c,\mathbf{y}_{<t}),
    \label{eq:unified_generation_en}
\end{equation}
Here, $T$ is the response length in tokens rather than a number of environment interaction steps, and $\mathbf{y}_{<t}$ denotes the preceding output tokens. The response may comprise a multiple-choice answer, a numerical answer, concise text, a coordinate sequence, or an ordered action plan. The five dimensions organize capability coverage rather than define separate training losses: we optimize causal cross-entropy over unmasked assistant tokens, normalized within each training sequence, and with a per-device microbatch size of one, gradient accumulation and data-parallel averaging combine these sequence-normalized losses across the effective batch. Inputs, media, and padding carry no supervision, and the assistant tokens of a multi-turn sample are normalized together. Relative exposure across capabilities is controlled by the data construction and sampling strategy of Section~\ref{sec:engine} rather than by per-task loss weights. At inference, all dimensions use this same generation procedure, without loading separate task-specific checkpoints and without an additional decoding stage.

This objective is constrained in practice by two obstacles: \textbf{Modality Entanglement}, under which heterogeneous supervision from the five dimensions competes over shared parameters, and \textbf{Data-Model Drift}, under which a fixed training-data mixture may become misaligned with the model's evolving learning needs. We first present the unified architecture that accommodates heterogeneous inputs and outputs, then introduce the Task-Conditioned Adaptive Training Strategy to mitigate interference between heterogeneous output types; Data-Model Drift is addressed by the Agentic Data-Model Self-Evolution Engine in Section~\ref{sec:engine}.

\subsection{Model Architecture}

To accommodate heterogeneous input modalities and output semantics within a single parameter space, ABot-ER adopts a unified decoder-only vision-language architecture based on the Transformer~\cite{vaswani2017attention} comprising three core components: a vision encoder $\mathcal{E}$, a modality projector $\mathcal{P}$, and an autoregressive language model $\mathcal{G}$. All five dimensions share the same autoregressive backbone and output head; task-conditioned residuals are introduced within selected attention projections rather than as separate per-dimension decoders. We instantiate this architecture on the pretrained multimodal model Qwen3.5-9B (approximately 9B parameters). Its language backbone comprises 32 decoder blocks with a hidden dimension of $d = 4096$ and a hybrid attention architecture: three Gated DeltaNet linear-attention blocks followed by one full-attention block, repeated eight times, so that the full-attention blocks are those with zero-based indices $3, 7, \ldots, 31$. When the task provides a visual observation $I$ (an image or a video frame sequence), the visual branch extracts $\mathbf{z}_v = \mathcal{P}(\mathcal{E}(I))$, whereas text-only tasks bypass the visual branch and require no placeholder visual input. The data loader injects the task conditioning token $c \in \{\texttt{[EQA]}, \texttt{[TASK\_PLAN]}\}$ into the content of the first user message, once per sample, without rewriting the frozen benchmark prompt body. That message is then rendered by the model's chat template, which supplies the role delimiters and an optional system message, and $\mathbf{z}_v$ is inserted at the media placeholders the template emits. Within the message, text and media keep their original interleaved order, so the sequence reaching $\mathcal{G}$ opens with the user role delimiter, then $c$, then the interleaved content: the task token sits inside the first user message rather than at the head of the input sequence. Writing $\mathbf{z}_{\mathrm{in}}$ for this input sequence, the output $\mathbf{y} = \mathcal{G}(\mathbf{z}_{\mathrm{in}})$ is generated autoregressively token by token, whether it comprises multiple-choice answers, coordinate points, natural language descriptions, or ordered action sequences. All dimensions use the same generation procedure without a separate rationale-decoding stage.

We map the five capability dimensions onto the two task families the training corpus is collected as, and condition on those families. Grouping at this granularity keeps conditioning coarse, with two branches rather than one per dimension, so that most parameters stay shared across all tasks. \texttt{[EQA]} covers embodied question answering, which supplies the spatial, scene-state and temporal supervision together with the interaction coordinates, and \texttt{[TASK\_PLAN]} covers task planning, which supplies ordered action plans. The families are defined by how the corpus was collected, not by the form of the answer: a coordinate is a question-answering record here, even though it is an action-space output. The two families share object, spatial, state, and affordance knowledge, but differ in supervision scale, answer length, output protocol, and reasoning horizon, and it is this difference that the conditional residuals are given capacity for. The route is determined from record-level task-family metadata and shared across all turns in a sequence; it does not enforce a uniform answer format across turns, and each family also carries auxiliary supervision in other answer forms. Because the task family of a query is known before inference, routing is a deterministic lookup rather than a learned router.

To remove format inconsistencies at the data level, ABot-ER separates the numeric protocol from the serialization protocol. The numeric coordinate convention is shared across tasks: coordinates are continuous values normalized to the $[0,1]$ interval relative to the original image dimensions, with three decimal places. This is the single coordinate convention the model has to learn. Serialization, by contrast, follows a fixed benchmark-specific format to comply with the corresponding evaluation interface. Section~\ref{sec:protocol} describes both layers and the conflict they resolve.

\subsection{Task-Conditioned Adaptive Training Strategy}
\label{sec:tcat}
 
The Task-Conditioned Adaptive Training Strategy augments the attention query and value projection modules of selected full-attention blocks with task-conditioned low-rank residuals, using a parameterization inspired by low-rank adaptation~\cite{hu2021lora}. Each residual operates on the corresponding projection input and is added to its output before subsequent attention computation. We place them in the last four full-attention blocks, whose zero-based indices form the set $\mathcal{S} = \{19, 23, 27, 31\}$. For $l \in \mathcal{S}$, a projection $p \in \{q_{\mathrm{proj}}, v_{\mathrm{proj}}\}$, and the hidden state $\mathbf{x}^l$ entering that projection:
\begin{equation}
    \widetilde{\mathbf{z}}^{\,l}_{p,c}
    =
    \mathbf{W}^l_p \mathbf{x}^l + \mathbf{b}^l_p
    +
    \frac{\alpha}{r}\, \mathbf{B}^l_{p,c} \mathbf{A}^l_{p,c} \mathbf{x}^l
    \label{eq:dhta_residual_en}
\end{equation}
where $\mathbf{A}^l_{p,c} \in \mathbb{R}^{r \times d}$, $\mathbf{B}^l_{p,c} \in \mathbb{R}^{d_{\mathrm{out},p} \times r}$, $r \ll d$, and $\alpha$ is a scaling coefficient. The output dimension $d_{\mathrm{out},p}$ follows the projection itself and therefore differs between the two. In this backbone $q_{\mathrm{proj}}$ jointly produces query features and attention output gates, which are split downstream; the residual is added to the full projection output before that split and therefore applies to both parts. These attention output gates are internal to the backbone and unrelated to the task routing described above.

Each projection holds one such pair of factors per route, so that $c$ selects between the two task-family sets. The route is resolved once per sequence and applies to all of its tokens. All parameters other than the route-specific low-rank factors are shared across tasks. The shared language backbone and the residuals are trained jointly, so the shared weights continue to receive gradients from all tasks; the residuals provide a task-family-specific adaptation path intended to mitigate interference between the two families, rather than isolating task gradients.

The rank $r$, the scaling coefficient $\alpha$, the residual layer indices, and the detailed per-stage training configuration are reported in Table~\ref{tab:training_config_en}. At inference time, the task marker selects the residual branch before response generation, and the selected branch remains active throughout decoding without requiring additional generated routing tokens. Section~\ref{sec:experiments} compares the task-conditioned configuration against an unconditional shared baseline, assessing task tokens and conditional residuals jointly rather than isolating the contribution of task tokens.

%% file: sections/data_agent_v3.tex
\section{Agentic Data-Model Self-Evolution Engine}
\label{sec:engine}

\begin{figure}[!htbp]
  \centering
  \includegraphics[width=\textwidth]{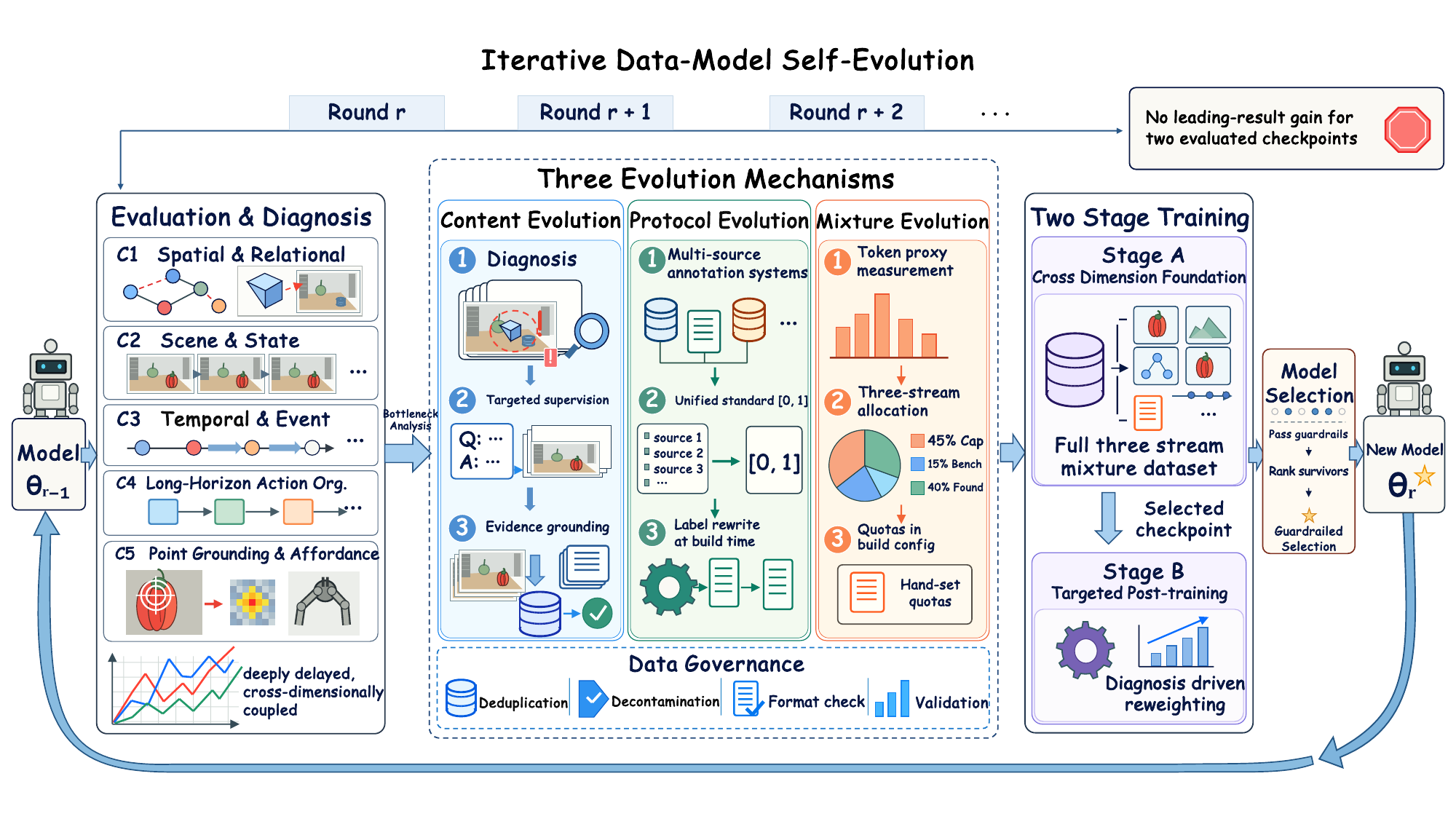}
  \caption{Closed-loop pipeline of the Agentic Data-Model Self-Evolution Engine. Each round begins with evaluation-driven capability diagnosis (Section~\ref{sec:diagnosis}), revises the training corpus through three evolution mechanisms, namely content (Section~\ref{sec:content}), protocol (Section~\ref{sec:protocol}), and mixture (Section~\ref{sec:mixture}), under a data governance layer (Section~\ref{sec:governance}), and closes with training and checkpoint selection (Section~\ref{sec:training}); the released model uses the two-stage recipe described there. The evaluation result of the new model version determines the revision applied in the next round.}
  \label{fig:data_model-circle}
\end{figure}

This section presents the Agentic Data-Model Self-Evolution Engine, an evaluation-driven framework that coordinates supervision content, output protocols, and data mixtures for joint embodied reasoning training. Its central design choice is to diagnose \emph{which aspect} of the training data requires revision, rather than treating every capability bottleneck as a demand for more samples. A capability can lag because the supervision it needs does not exist, because that supervision exists but carries mutually exclusive targets, or because correct targets receive too little budget to take effect. These call for different repairs, and one diagnosis often calls for more than one of them.

The need for revision is continuous rather than preparatory. As the model's capability profile changes, the supervision and exposure needed for further improvement change with it, so a mixture selected at the outset may become less effective in later rounds. A revision must also be assessed through subsequent training and evaluation across five dimensions and twelve benchmarks, since gains in one dimension may accompany regressions in another. Related work explores synthetic instruction generation, iterative self-feedback and self-rewarding, and data-mixture optimization~\cite{wang2022selfinstruct,yuan2024selfrewarding,madaan2023selfrefine,xie2023doremi,gadre2023datacomp}. Here the focus is on using a shared capability diagnosis to coordinate supervision content, task-specific output protocols, and exposure allocation in joint embodied reasoning training.

The following subsections describe the shared diagnosis (Section~\ref{sec:diagnosis}), the three coordinated revisions (Sections~\ref{sec:content}--\ref{sec:mixture}), and the governance and training procedures used to evaluate their outcomes across rounds (Sections~\ref{sec:governance}--\ref{sec:rounds}). Full construction quantities, audits, configurations, and the per-round ledger are collected in Appendix~\ref{app:engine}.

\subsection{Evaluation-Driven Capability Diagnosis}
\label{sec:diagnosis}

Diagnosis decides what the next corpus revision should change. A round freezes the evaluation contract so that scores are comparable across its checkpoints, aggregates results at the granularity of the five capability dimensions, classifies the underlying errors, and passes the resulting gap profile to the coordinating agent, which returns a data requirement together with candidate sources.

The contract fixes the prompt text, media renderer, output parser, and scorer of every benchmark, together with the chat template and decoding mode, and is identified by a content hash. It provides a stable target for training-label alignment, which label construction and parser round-trip checks then enforce. Any change to a prompt, a required point count, a parser, or a scorer produces a new contract version, and scores obtained under different versions are not treated as directly comparable without re-evaluation under a common contract. Because biases introduced early in a closed loop compound over later rounds, the initial rounds are human-calibrated: an expert reviews the agent's diagnoses and the resulting revisions before they enter training, and those judgements are encoded as the deterministic rules and gates that later rounds run against.

\subsubsection{From Symptom to Fault}

The choice of diagnostic granularity is consequential. Using a single benchmark's score as the evolution signal carries a risk associated with Goodhart's Law~\cite{manheim2018goodhart}: when a metric is continuously optimized, it may cease to faithfully reflect the capability it was designed to measure. Targeted augmentation for a specific benchmark may teach the model that benchmark's idiosyncratic answering patterns, improving the score without strengthening spatial reasoning. We therefore separate the \emph{symptom}, which is necessarily observed at benchmark level, from the \emph{conclusion}, which is drawn at capability-dimension level. Dimension scores are fixed equal-weight averages over the benchmark groups of Table~\ref{tab:metrics_en}. A capability gap is admitted only when the benchmarks within a dimension exhibit consistent trends, or when a benchmark-specific symptom is independently traced to a mechanical cause such as a protocol violation. For a dimension represented by a single benchmark, cross-benchmark agreement is unavailable, and diagnosis relies on item-level error analysis instead. Which of these applies is decided by the error classification, not by inspection of the scores alone.

Raw model outputs and error logs are classified by fixed rules into six classes, each read off the record itself rather than inferred from the score: parser and format failures; point misses separated from coordinate protocol violations; multiple-choice category errors; numeric parsing errors; open-ended judge deductions; and task-plan errors, subdivided into illegal actions, truncated plans, and executable plans that fail to reach the goal state. These classes are then mapped onto the five dimensions to obtain an attributed gap profile. Table~\ref{tab:diagnosis_rules_en} lists representative mappings actually used, each with the check that separates a capability conclusion from a benchmark-specific artifact; where a symptom is traced to the contract, no such check is needed. Deterministic checks classify observable errors and establish audit findings. The coordinating agent uses this evidence to propose root-cause hypotheses and corresponding data requirements.

The fault categories organize intervention targets. A diagnosis may identify a primary intervention while requiring coordinated revisions to supervision content, protocol, and exposure: content evolution constructs or repairs the required targets, protocol evolution aligns their representation with task-specific interfaces, and mixture evolution adjusts how much exposure they receive. Increasing exposure does not repair incorrect targets or incompatible output conventions, while valid supervision may remain ineffective when insufficiently sampled. Two of the five categories concern evaluation-interface compliance and evaluation integrity instead. A \emph{contract} fault indicates non-compliance with the frozen evaluation interface; output and training-label checks determine whether the violation originates in the labels or occurs despite compliant supervision, and the labels are rebuilt, without adding samples, only when a training-label mismatch is confirmed. A \emph{contaminated evaluation} means the score no longer reflects capability; an anomalous score triggers a contamination audit and is classified this way only when the audit confirms overlap, after which the implicated sources are quarantined (Section~\ref{sec:governance}). The other three place the intervention on the supervision. \emph{Missing} or incomplete targets primarily motivate content construction (Section~\ref{sec:content}). \emph{Contradictory} supervision may require content repair when target semantics disagree, protocol alignment when representations disagree (Section~\ref{sec:protocol}), or both. \emph{Under-exposure} motivates mixture reallocation under the relevant content and protocol checks (Section~\ref{sec:mixture}). Section~\ref{sec:mixture} traces a case in which quota adjustment was followed by persistent supervision conflicts, motivating further content and protocol revisions.

\begin{table}[!htbp]
\centering
\caption{Representative mappings from evaluation symptom to attributed fault and data action. The symptom triggers the investigation, and the diagnostic check is the evidence available when the fault is assigned: contaminated evaluation and contradictory supervision are confirmed by the audits named there. The fault class determines which mechanism primarily handles the symptom.}
\label{tab:diagnosis_rules_en}
\small
\renewcommand{\arraystretch}{1.15}
\setlength{\tabcolsep}{3pt}
\begin{tabular}{p{0.19\linewidth}p{0.18\linewidth}p{0.14\linewidth}p{0.36\linewidth}}
\toprule
\textbf{Symptom} & \textbf{Diagnostic check} & \textbf{Fault} & \textbf{Data action} \\
\midrule
Unparseable or off-contract outputs & Output and training labels checked against the frozen renderer and parser & Contract & Labels rebuilt against the frozen renderer and parser when a label mismatch is confirmed; no new samples added \\
\addlinespace[2pt]
Anomalously high EB-ALFRED & Source-level audit confirms overlap with the official evaluation trajectories & Contaminated evaluation & Source-level leakage audit and quarantine of the implicated sources \\
\addlinespace[2pt]
Low EB-ALFRED and EB-Habitat with legal actions & ET-Plan consistent & Missing supervision & Replay-verified multi-step plans with explicit precondition chains and verified terminal states \\
\addlinespace[2pt]
Low Cosmos & Item-level analysis: errors concentrate on temporal-order and failure-recovery items rather than on static-content items & Missing supervision & Train-only video multiple choice balanced across five independent clean domains \\
\addlinespace[2pt]
Low VABench-P with well-formed 8-point output & Where2Place moves in the same direction; training-label inspection identifies incompatible target conventions & Contradictory supervision & Point supervision restricted to verified target-region labels; trajectory endpoints removed \\
\addlinespace[2pt]
A dimension lags while its benchmarks agree and its supervision is already verified & Per-source and per-group exposure counts confirm the dimension is under budget & Under-exposed supervision & Dimension quota raised under the source and exposure caps; no change to the labels themselves \\
\bottomrule
\end{tabular}
\end{table}

\subsubsection{Agent Proposals and Deterministic Verification}

What the rules do not supply is the hypothesis behind a gap and the candidate sources that would close it. That open-ended step, reading a score curve together with per-item errors and a source profile and proposing a root cause, is performed by a single coordinating language-model agent, realized as an interactive coding agent. Evaluation itself runs in an execution context separate from that agent, which separates score production from corpus revision under the frozen evaluation contract. The agent's inputs are deliberately asymmetric across the two halves of its work. For \emph{diagnosis} it reads the evaluation records, per-sample outputs, label and media metadata, training logs, and audit reports. For \emph{construction}, evaluation items are not supplied as source material: a new label is written against a small set of format-reference exemplars, disjoint from every evaluation set, that fix the output contract it must satisfy. Its outputs are a diagnosis and a structured data request: a root-cause hypothesis, the supervision to add, the sources to remove, and the guardrails the resulting build must pass.

Responsibility is partitioned so that the agent proposes and programs verify. The coordinating agent proposes root-cause hypotheses, data requirements, and candidate sources. Deterministic programs verify each candidate build against the applicable format, provenance, exposure, and replay checks, and no proposal enters the formal training pool without passing the gates specified for that release.

Two archived records show what this produces. Case~B, a missing-supervision fault, drove the replay-verified planning corpus and is traced in Section~\ref{sec:content}; Case~A, which mixes a target-semantics problem with a protocol conflict, drove the point-supervision rebuild and is traced in Section~\ref{sec:protocol}. Under-exposure is traced in Section~\ref{sec:mixture} and the contaminated-evaluation fault in Section~\ref{sec:case_c}.

\subsection{Content Evolution: Targeted Supervision}
\label{sec:content}

Content evolution turns a diagnosed failure into a specification of the supervision to construct or repair. For point grounding, this means targets tied to verified task-relevant regions rather than motion endpoints. For planning, it means complete trajectories with checked preconditions and verified terminal states. The aim is to supply the missing or corrected learning signal, not merely to enlarge the corpus.

Case~B shows how a failure becomes such a specification. In an archived record from the R3 diagnostic round, six actions succeeded without completing the task, and later planner calls returned unparseable output: the initial plan omitted the placement action, while a repetitive state description left the subsequent JSON unclosed. These observations motivated requirements for terminal-state-complete trajectories verified by simulator replay, feedback-conditioned recovery examples, bounded state descriptions, and parser-valid labels. Table~\ref{tab:supervision_methods_en} summarizes the construction method used for each dimension.

Two risks accompany targeted construction. The first is \textbf{supervision reliability}: agent-assisted supervision may contain unverified fabrications that would inject systematic noise. We therefore adopt evidence grounding, so that every training target is derived from input media, environment state records, or trusted annotations, with the agent acting only as a candidate proposer and its outputs entering the formal pool only after independent verification. Deterministic release audits then enforce the specified constraints on format, provenance, coverage, and exposure that every build must pass. The second is \textbf{diminishing marginal returns}, which made the acceptance criteria themselves a target of revision. A construction requirement stated as an exact count makes the last few percent of a target as expensive as the first ninety, because the items still missing are the ones the available sources cover least well, and a build can spend far longer closing that remainder than it spent producing the bulk. Construction targets are therefore specified as an elastic band rather than an exact count, and the engine tracks the cost of each additional accepted item as the round proceeds. When the measured construction difficulty for the remainder is high, the band widens and the round concludes on a documented exception, provided the residual gap does not materially alter capability coverage or risk levels.

\begin{table}[!htbp]
\centering
\caption{Supervision construction by capability dimension. Released record counts, historical processing volumes, and the exact label recipes are reported in Appendix~\ref{app:engine}.}
\label{tab:supervision_methods_en}
\small
\renewcommand{\arraystretch}{1.15}
\setlength{\tabcolsep}{5pt}
\begin{tabular}{@{}p{0.26\textwidth}p{0.68\textwidth}@{}}
\toprule
\textbf{Capability dimension} & \textbf{Supervision construction} \\
\midrule
Spatial and relational & Train-split spatial annotation, cleaned, relabelled, and sampled for coverage across fine-grained capability channels \\
\addlinespace[2pt]
Scene and state & Existing question-answer pairs cleaned of answer prefixes and markup, together with indoor question answering constructed from replay-passed scene and state records \\
\addlinespace[2pt]
Temporal and event & Video items that retain the temporal evidence, with determinate answer conversion and sampling across independent clean train domains \\
\addlinespace[2pt]
Point grounding and affordance & Supervision restricted to verified target regions, serialized to the point cardinality each task requires \\
\addlinespace[2pt]
Long-horizon action organization & Planning trajectories with checked preconditions and verified terminal states, together with feedback-conditioned recovery windows \\
\bottomrule
\end{tabular}
\end{table}

At the level of supervision structure, the five dimensions share foundational representations of objects, spatial relations, states, and affordances, but retain distinct answer formats. Perception retains multiple choice, numeric values, and concise text, emphasizing evidence-grounded state readout; planning retains complete ordered action plans, emphasizing sub-goal dependencies, action ordering, and cross-step state consistency, and execution retains spatial coordinate outputs for task-relevant interaction targets. Unification applies to coordinate values and metadata, not to capability-specific output forms.

\FloatBarrier
\subsection{Protocol Evolution}
\label{sec:protocol}

Protocol evolution revises inconsistent training representations to satisfy task-specific output requirements. It unifies the numeric coordinate convention while preserving the point cardinalities, serialization formats, and action schemas that different tasks require. Training data for different dimensions often inherits from independently developed annotation systems, producing incompatible representations of the same underlying quantity. For point supervision, three conventions were in simultaneous use: legacy supplementary data using a $0$--$1000$ coordinate range with eight points wrapped in XML or JSON; historical training rewrites using a single $[0,1]$ tuple \texttt{[(x,y)]}; and the official evaluator, which requires $[0,1]$ coordinates, exactly eight points, and the XML form \texttt{<Answer><point>[[x1,y1], \ldots, [x8,y8]]</point></Answer>}. With conflicting targets acting on shared parameters for the same image-question pairs, the model has no single output convention to fit.

Case~A shows why the two are separable. In an archived point-grounding record, all eight predicted points were parseable and contract-compliant, yet none fell inside the target region: a representation that satisfies the interface can still carry the wrong target. Historical supervision mixed target medoids, motion endpoints, and arm or hand paths, while the training-data audit separately identified the incompatible ranges, counts, and wrappers above. Before the combined revision, VABench-P scores spanned a range of $45.17$ points across the evaluated checkpoints of one run. The rebuild retained verified target-region supervision and aligned its representation with each task's requirements, after which the within-run range narrowed to $3.46$ points, an order of magnitude tighter. On the format-aligned corpus the released model trains on, two runs under one configuration hold within-run spreads of $2.33$ and $3.00$ points once past the early training ramp, and agree to within $1.67$ points at matched checkpoints.

\begin{table}[!htbp]
\centering
\caption{VABench-P around the coordinate-protocol unification, and its reproduction on the released corpus. The upper block is an earlier development line, where point supervision carried three coordinate conventions at once; its scores are development history, not results of the released model. The lower block is the format-aligned corpus the released model trains on, run twice under one configuration. Spread is the range over the later evaluated checkpoints of each run, excluding the early training ramp; over that range the two runs differ by at most $1.67$ points at matched checkpoints.}
\label{tab:vabench_protocol_en}
\small
\renewcommand{\arraystretch}{1.15}
\setlength{\tabcolsep}{6pt}
\begin{tabular}{lcc}
\toprule
 & \textbf{VABench-P} & \textbf{Within-run spread} \\
\midrule
\multicolumn{3}{l}{\textit{Development history, earlier line}} \\
\addlinespace[2pt]
Pretrained model & 24.85 & --- \\
Before unification & 8.58--53.75 & 45.17 \\
After unification and rebuild & 64.54--68.00 & 3.46 \\
\addlinespace[5pt]
\multicolumn{3}{l}{\textit{Released corpus}} \\
\addlinespace[2pt]
Same configuration, run 1 & 61.67--64.00 & 2.33 \\
Same configuration, run 2 & 61.33--64.33 & 3.00 \\
\midrule
Released checkpoint & \textbf{63.33} & --- \\
\bottomrule
\end{tabular}
\end{table}

The resolution separates the two layers. The numeric convention is unified: all spatial coordinates in training, across all dimensions, are continuous values normalized to $[0,1]$ relative to the original image dimensions, with three decimal places, and this is the single coordinate convention the model has to learn. Legacy labels on the $0$--$1000$ range were rescaled into that convention rather than discarded, so that no source was dropped for its coordinate range alone. Bounding-box supervision was removed instead, and at turn granularity: turns whose assistant output is a box left the training pool, while samples that merely carry a box in their input and answer in text or points were retained. Serialization, by contrast, stays benchmark-specific and is frozen into the training labels against each evaluation contract, preserving the point cardinality, wrapper format, and action schema that contract requires: eight XML points for VABench-P, a one-to-three-point tuple list for Where2Place, a single A--D token for the multiple-choice benchmarks, concise native text for the open-ended ones, and their own four-field JSON and action spaces for ALFRED, Habitat, and ET-Plan. Task routing enters through the loader as a prefix token and never rewrites the frozen benchmark prompt body.

Build-time audits provide check-specific evidence for this design. On the released post-training stage every one of its $104{,}960$ labels passes the contract check; on the pre-training stage all $46{,}776$ VABench-P labels are valid eight-point XML. Both stages record zero bounding-box outputs and zero non-canonical points.

\subsection{Mixture Evolution}
\label{sec:mixture}

Mixture evolution adjusts exposure for supervision whose content and output requirements have already been checked. It addresses insufficient sampling without treating reweighting as a substitute for correcting targets or protocols. The point-grounding history illustrates the distinction: quota adjustment was followed by a higher dimension score, but substantial checkpoint variation remained until the content and protocol revisions were also applied. Sample counts alone do not capture differences in input size or supervision length. During corpus construction the system therefore uses a build-time token proxy as a common budget metric, computed from the characters of all conversation-message contents together with per-image and per-video allowances; it is a character heuristic rather than a tokenizer count.

No single accounting basis is sufficient, and the three used in this report answer different questions: \emph{unique and physical records} reveal repeated exposure and source coverage; \emph{consumed samples}, equal to optimizer steps times global batch, reveal how much data the model actually saw and govern streaming shard safety; \emph{assistant supervision tokens}, measured with the model's own tokenizer, quantify supervised output volume. Because the objective is normalized within each training sequence (Section~\ref{sec:architecture}), a supervision-token share describes the corpus rather than a proportional weight in the training objective. The bases can diverge sharply. In R6, the single-trunk development release of Table~\ref{tab:round_scores_en} rather than either stage of the released model, the question-answering-to-planning ratio is $75\!:\!25$ by unique records but $29.71\!:\!70.29$ by assistant supervision tokens, because planning labels are long structured plans and many question-answering labels contain only a single token. Every mixture figure in this report therefore names its basis.

In R6, the question-answering partition is allocated over unique records in three streams. A fine-grained capability stream, targeted at approximately $45\%$, distributes budget equally across fine-grained capability channels so that training contribution is decoupled from raw volume. A benchmark guarantee stream, approximately $15\%$, sets a minimum quota for each of the nine question-answering benchmarks so that none can be starved by the fine-grained allocation; its budget is filled by drawing train-split supervision from those benchmarks, and is reweighted toward the dimension that diagnosis identifies as marginally weak. A foundational capability stream, approximately $40\%$, covers four general knowledge categories in equal proportion so that the model retains broad competence beyond evaluation-specific distributions. Planning supervision sits outside these three streams and is allocated as separate replay-verified trajectory pools.

To prevent degenerate allocations, boundary constraints are imposed on the question-answering mixture. Per-benchmark and per-source shares are bounded within the question-answering token-proxy budget, counting materialized exposures so that repetitions accumulate, and no sample is exposed more than four times. The numbers themselves are set by hand: target quotas per dimension and per benchmark are read off the evaluation diagnosis of the round and recorded as literals in its build configuration, while the coordinating agent supplies the qualitative data requirement and the candidate sources rather than the quotas. Deterministic programs then validate those quotas against the bounds above and realize them by capped stratified sampling, drawing samples stratified by source and by group under the exposure caps, where all variants of the same scene, episode, or trial share one group budget; within the manually fixed stream budgets, the pretraining mixture allocates the benchmark-by-source cells with a linear program. If any capacity or format gate is unsatisfied, the build fails and is rolled back. Each release records an ordered SHA-256 over its serialized records, and rebuilding from the same seed reproduces that hash. Learned mixture optimization~\cite{xie2023doremi} and influence-based data selection~\cite{xia2024less} address the same allocation problem with automatic criteria and represent a natural direction for replacing the manually set quotas used here.

The point grounding and affordance dimension is the engine's clearest under-exposure fault, and also its clearest illustration of why exposure adjustment alone is insufficient when supervision conflicts remain. In the earliest round the dimension took whatever share source frequency gave it and scored $50.58$. The first build with explicit quota constraints raised the two point benchmarks to $15.04\%$ of the total token proxy, and the dimension scored $57.39$ in that round. That quota-adjusted build nevertheless showed substantial checkpoint variation on VABench-P, with a within-run range of $45.17$ points (Table~\ref{tab:vabench_protocol_en}). Following protocol alignment and the Case~A target-semantics rebuild, the reported VABench-P range narrowed to $3.46$ points. The released model reached a point grounding and affordance dimension score of $67.06$, with Stage~B allocating $25.53\%$ of its physical records and $29.96\%$ of its assistant supervision tokens to point supervision. The ordering is the lesson: the dimension carried an under-exposure fault and a contradictory-supervision fault at once, and quota decides only the first of the two.

\FloatBarrier
\subsection{Data Governance and Evaluation Integrity}
\label{sec:governance}

Data governance protects the evidence the evolution loop runs on. Each build is checked for source eligibility, supervision consistency, media integrity, and overlap with evaluation data under the specified audit rules. Deduplication proceeds at three levels: intra-source exact identity matching; cross-source media and question hashing, which catches the same annotation set mounted under two source names; and quarantine of conflicting answers for the same image and question, adjudicated rather than silently discarded. The four audited risks are the following.

\begin{itemize}
    \item \textbf{Evaluation contamination.} If training data covers test content, evaluation loses diagnostic validity~\cite{oren2023contamination} and the entire feedback loop is corrupted. Two controls apply within the loop (Section~\ref{sec:diagnosis}): evaluation runs in an execution context separate from the coordinating agent, and evaluation items are excluded as source material for supervision construction. Leakage detection then screens at instruction, trajectory, and environment granularity. Even absent direct answer overlap, training data that covers all test environments and supervises environment-specific identifiers is flagged as a potential risk.
    \item \textbf{Format interference.} Invalid coordinate values or labels inject spurious gradient signal; format compliance verification intercepts such samples before they enter the training pool.
    \item \textbf{Reference integrity.} Broken media references cause the model to receive blank visual input; media integrity verification requires every record to resolve to an existing file.
    \item \textbf{Semantic and provenance compliance.} Sources are checked against eligibility and licensing constraints, and semantic near-duplicates and anomalous perfect scores are escalated for case-by-case adjudication against the criteria calibrated in the initial rounds rather than resolved by threshold alone.
\end{itemize}

\paragraph{Case~C: a contaminated-evaluation fault.}\label{sec:case_c}
The fourth criterion is not a formality, and it is the one that catches the fault no capability fix would have addressed. In round R3 the reported EB-ALFRED score was $100$. Because the diagnosis operates at dimension level, this was read not as a capability result but as a within-dimension inconsistency: ET-Plan and EB-Habitat showed no comparable movement, which triggered the contamination audit of Table~\ref{tab:diagnosis_rules_en}. The audit found two independent defects. The score itself was an aggregation artifact: the run summary had been written while only $12$ of the $300$ episode results existed, all of which happened to succeed, and a later stage reused that archived summary instead of recomputing over the full set. Recomputing over all $300$ episode results, on the same checkpoint and under the same contract with no further inference, gives $27$. Separately, a source-level audit confirmed that the legacy task-planning pool contained official EB-ALFRED test trajectories, supervising the official instruction text and per-step image paths. Neither number is therefore usable as an EB-ALFRED result for that checkpoint: the $100$ never measured capability, and the recomputed $27$ was produced by contaminated weights. Evaluation completeness became an acceptance criterion in response, requiring a summary to cover the full episode set and disallowing reuse of an archived summary.

The remediation was source-level rather than sample-level. Three implicated trajectory sources were removed in their entirety when building the next release, quarantining $23{,}565$ records at build time rather than filtering them after training. The quarantine applies to the candidate pool rather than to the released records, which are resampled under the same quotas, so the released record count is close to that of the preceding round. R3 is retained in this report as a diagnostic round only: its scores are reported in Table~\ref{tab:round_scores_en} for completeness and are excluded from every public performance claim.

Acceptance criteria are partitioned into hard gates, which must pass, and soft targets, which permit bounded and documented deviation. Hard gates are adjudicated against the rule set in force for that round, and a dataset is registered as a trainable release only after all of them pass; Appendix~\ref{app:engine} states them together with the result recorded for each.

\subsection{Training, Selection, and Evolution Across Rounds}
\label{sec:training}
\label{sec:rounds}

This subsection describes the two-stage recipe used for the released model; earlier development rounds used their own configurations, documented in Appendix~\ref{app:engine}. Both stages train on corpora this engine produced: Stage~A on the release that the content, protocol, and mixture revisions of the earlier rounds converged on, and Stage~B on the point-heavy release specified by the Case~A rebuild together with the under-exposure diagnosis of Section~\ref{sec:mixture}. Training first establishes competence across the joint capability mixture, then reallocates supervision toward the remaining bottlenecks while retaining other-task data to mitigate forgetting; the selected checkpoint, rather than the final training step, is carried forward. The two-stage design is motivated by asymmetric convergence observed in joint training. \textbf{Stage~A}, cross-dimension foundation, starts from the pretrained vision-language model and trains on the joint mixture, comprising the three-stream question-answering partition and the separately allocated planning supervision, with a large budget. Its objective is to establish baseline competence across all five dimensions, internalize the unified numeric protocol and the frozen per-benchmark serializations, and learn task-conditioned output switching. After sufficient iterations Stage~A exhibits asymmetric convergence: perception performance saturates first while planning and execution continue improving. Planning and execution require longer exposure to establish stable generation patterns for complete action sequences and precise coordinates.

This asymmetric progress motivates \textbf{Stage~B}, targeted post-training. It is supervised fine-tuning rather than reinforcement learning, and proceeds from the selected Stage~A checkpoint on a set rebalanced toward the bottleneck dimensions. Because concentrated training risks catastrophic forgetting, it uses a substantially reduced learning rate, retains a proportion of saturated-dimension data as anti-forgetting anchors, and runs for a much shorter schedule.

The released model uses $462{,}390$ physical records for Stage~A, split $90.44\!:\!9.56$ by physical records between question answering and planning, and $104{,}960$ physical records for Stage~B, of which point supervision accounts for $25.53\%$ by physical records and $29.96\%$ by assistant supervision tokens. Stage~B reduces the decoder learning rate from $4.0\times10^{-6}$ to $3.0\times10^{-7}$ while retaining other-task data as anti-forgetting anchors. In each stage an earlier checkpoint is carried forward rather than the final one. The complete jobs consume approximately $826$ GPU hours on 32 GPUs, excluding data construction and full-suite evaluation, which is what makes a revision expensive to assess and worth diagnosing before it is made. Full compositions and configurations are reported in Appendix~\ref{app:engine}.

\paragraph{Checkpoint selection.} Model selection evaluates several checkpoints of a run under the frozen contract, filters them by guardrails, and ranks the survivors in a fixed order. A guardrail is a benchmark-level retention bound rather than a capability dimension: each round fixes a floor on the nine-benchmark question-answering mean, bounds on the permitted drop of the task-planning mean and of the individual benchmarks the round is not targeting, a parser-reliability floor, and zero detected overlap with evaluation data. These bounds are stated per round against the checkpoint the run started from, so no single tolerance applies across the history; those in force for the released model are listed in Appendix~\ref{app:engine}. Among the checkpoints that clear them, selection compares in order the number of leading results at that checkpoint, then the mean over the benchmarks the round targets, then the nine-benchmark question-answering mean. A leading result is a benchmark on which the checkpoint strictly exceeds the best score of a frozen set of public models, counted within a single checkpoint so that a profile assembled from several checkpoints cannot stand in for one model. The peak of any single dimension is not used as the criterion, because in joint learning an extreme score on one dimension typically accompanies significant degradation elsewhere. Targets and guardrails together span all twelve benchmarks, so no capability dimension goes unwatched: the released run maximizes a benchmark in four of the five dimensions and bounds every remaining benchmark against regression, with long-horizon action organization covered by bounds on the planning mean and on each of its three benchmarks.

\paragraph{Retrospective selection and round closure.} The engine applies no fixed numerical convergence threshold. Each run is trained to its configured length, and the selection window is determined retrospectively from the evaluated checkpoints: a round is closed, and its checkpoint either selected or rolled back, when two successively evaluated checkpoints fail to increase the number of leading results on the targeted benchmarks and either the targeted mean declines or a guardrail is breached. When that condition does not arise, the round closes once the configured budget is exhausted, and selection proceeds over all checkpoints evaluated in the run. When in addition the remaining bottlenecks cannot be addressed by revising the mixture or the source pool, the data version is frozen and the release proceeds to comprehensive pre-release evaluation. Figures~\ref{fig:capability_evolution_summary_en} and~\ref{fig:evolution_trajectories_en} in Section~\ref{sec:experiments} report the five capability summaries at the selected observation points of this history.

\paragraph{Cross-round continuation.} The development history is not a single continuation chain. Revisions that changed the coordinate protocol, the source eligibility rules, or the grounded-planning supervision were retrained from the pretrained model, since prior-protocol knowledge creates interfering priors; incremental reweightings continued from an earlier checkpoint. The R6 development runs of Table~\ref{tab:round_scores_en} resumed from the R4 checkpoint, deliberately bypassing the fragmented R5 weights. The released model instead followed the two-stage recipe above: Stage~A was initialized from the pretrained model on the open-source release derived from R4, and Stage~B from the selected Stage~A checkpoint. Since R4 itself restarted from the pretrained model after the implicated sources had been removed, no weights trained on those sources were carried into either branch. All continuations reset optimizer and scheduler state. The complete lineage is documented in Appendix~\ref{app:engine}.

\paragraph{Across rounds.} Table~\ref{tab:round_scores_en} reports the five-dimension profile of each round; the corresponding triggers, data revisions, and weight origins are recorded in Appendix~\ref{app:engine}. Rounds are numbered R1 to R6 in the order they were run. Two of them carry a setting worth naming here: R4 is the round rebuilt after the contaminated sources of Section~\ref{sec:case_c} had been removed from the candidate pool, and R6 consolidates the earlier fragmented repairs into a single release trained in one uninterrupted optimizer run. The source of gains changes structurally across the history, and round-to-round change is not monotonic: improvement on one dimension is frequently accompanied by regression on another.

\begin{table}[!htbp]
\centering
\caption{Five-dimension profile at a representative checkpoint of each round. R3 is reported for completeness only and is excluded from all performance claims, since its EB-ALFRED score is affected by the contamination of Section~\ref{sec:case_c}.}
\label{tab:round_scores_en}
\small
\renewcommand{\arraystretch}{1.12}
\setlength{\tabcolsep}{3.5pt}
\fitwidth{%
\begin{tabular}{lrrrrr}
\toprule
\textbf{Round} & \textbf{Spa.} & \textbf{Sce.} & \textbf{Tem.} & \textbf{Poi.} & \textbf{Lon.} \\
\midrule
Base & 35.26 & 46.45 & 51.64 & 35.25 & 33.07 \\
R1 & 62.32 & 60.13 & 62.87 & 50.58 & 35.23 \\
R2 & 61.41 & 57.48 & 61.87 & 57.39 & 32.98 \\
R3 & 62.70 & 60.89 & 66.56 & 42.15 & 41.86 \\
R4 & 62.11 & 61.88 & 68.64 & 47.80 & 42.96 \\
R5 & 62.88 & 62.21 & 69.16 & 51.73 & 46.89 \\
R6 & 63.42 & 62.29 & 67.75 & 66.13 & 45.57 \\
\midrule
ABot-ER (released) & \textbf{64.10} & \textbf{61.93} & \textbf{68.96} & \textbf{67.06} & \textbf{41.61} \\
\bottomrule
\end{tabular}%
}
\par\smallskip
\parbox{\linewidth}{\footnotesize
Spa.: spatial and relational. Sce.: scene and state. Tem.: temporal and event. Poi.: point grounding and affordance. Lon.: long-horizon action organization. Each row is reported under the evaluation contract in force for its round; when the contract changed, re-evaluation extended back over the two most recent trainings, so rows further apart in the table may sit under different contract versions.}
\end{table}

Across the history the source of gains moves through three phases. Early rounds are dominated by \textbf{structural corrections}: protocol alignment removes incompatible output conventions, contamination removal restores the interpretability of evaluation, and elementary mixture balancing gives each dimension a viable budget. Across these rounds the engine successively revised exposure allocation, target semantics, and output protocols for point grounding and affordance, and the dimension's reported score moves from the mid-thirties to the mid-sixties over that span. Those rows are separated by contract changes as well as by simultaneous revisions to training budget, sampling balance, coordinate conventions, and data quality. Middle rounds are dominated by \textbf{formalization of the procedure}: per-round tuning is replaced by explicit quotas, caps, and reproducible sampling with an ordered release hash, quality gates intercept defects before training rather than after, and the evaluation contract is frozen and versioned so that later comparisons mean something. Later rounds yield \textbf{targeted, single-digit improvements confined to specific dimensions}, at the price of regressions elsewhere: relative to R6, the released model is $0.93$ points higher in point grounding and affordance and $3.96$ points lower in long-horizon action organization. These are dimension-specific trade-offs, at a stage where the loop is no longer discovering structural defects.

\FloatBarrier

\paragraph{One pass or many.} A sequence of post-training rounds and the corpus it converges on are two different things, and it is worth asking which of them carries the benefit. We compare a single post-training pass against a chain of six. The two conditions share a common ancestor checkpoint and use the same corpus for the final post-training pass, but differ in prior training history and in how much post-training each has accumulated: the chain has had five times as much, and the two are compared at the same point within their common final pass. Table~\ref{tab:onepass_en} reports both. The single pass achieves a similar twelve-benchmark mean with a fifth of the accumulated post-training, but does not reproduce the same capability profile: it is ahead on question answering and on single-shot planning, and behind on the two interactive planning benchmarks. This supports direct reuse of the converged corpus. Section~\ref{sec:scaling} takes this at its word and trains four further backbones, from 2B to 27B parameters, with one pass each on the same two releases.

\begin{table}[!htbp]
\centering
\caption{A single post-training pass against a chain of six. Both start from the same ancestor checkpoint and use the same corpus for their final pass, and are compared at the same point within it; the chain has accumulated five times as much post-training.}
\label{tab:onepass_en}
\small
\renewcommand{\arraystretch}{1.15}
\setlength{\tabcolsep}{4pt}
\fitwidth{%
\begin{tabular}{lrrrrr}
\toprule
 & \textbf{EQA(9)} & \textbf{ET} & \textbf{ALF} & \textbf{Hab} & \textbf{All(12)} \\
\midrule
Single pass & \textbf{64.44} & \textbf{66.34} & 25.00 & 48.67 & 60.00 \\
Six rounds & 63.98 & 63.52 & \textbf{30.00} & \textbf{51.67} & \textbf{60.09} \\
\bottomrule
\end{tabular}%
}
\end{table}

\FloatBarrier

%% file: sections/er-experiment.tex
\section{Experiments}
\label{sec:experiments}

We first assess the overall competitiveness of a unified embodied model across twelve benchmarks and five capability dimensions, then apply the same training recipe to backbones from 2B to 27B parameters, followed by analyses of the Task-Conditioned Adaptive Training Strategy and data-model co-evolution. We examine whether competitiveness across tasks coexists within one model, whether task conditioning is associated with cross-task gains, and how iterative updates balance improvement with retention. Real-world demonstrations then illustrate the utility of model outputs within robotic systems.

\subsection{Experimental Setup}

\subsubsection{Capability Dimensions and Benchmarks}

We organize twelve benchmarks by the five dimensions introduced in Section~\ref{sec:architecture}. Perception comprises spatial and relational reasoning, scene and state understanding, and temporal and event reasoning. Planning is assessed through long-horizon action organization, while execution-oriented evaluation focuses on point grounding and affordance prediction. The latter measures spatial grounding for manipulation and does not, by itself, measure physical task completion. The nine-benchmark Embodied Question Answering (EQA) suite and the three-benchmark planning suite are evaluation collections rather than a partition into the three core competencies: EQA includes both perceptual judgments and execution-oriented pointing tasks.

\paragraph{Spatial and Relational Reasoning.}
EmbSpatial~\cite{du2024embspatial}, VSI~\cite{yang2024vsibench}, and ERQA~\cite{gemini2025robotics} provide complementary assessments of spatial understanding. EmbSpatial focuses on spatial relations from embodied viewpoints. VSI tests spatial layout, distance, size, and related reasoning from video observations. ERQA (Embodied Reasoning Question Answering) evaluates embodied reasoning through multiple-choice questions with interleaved images and text, including spatial and multi-view understanding. Its placement in this group indicates the principal emphasis of our analysis rather than an exhaustive characterization of its question types.

\paragraph{Scene and State Understanding.}
OpenEQA~\cite{majumdar2024openeqa}, UniEQA~\cite{unieqa_repository}, and RoboVQA~\cite{sermanet2023robovqa} assess whether a model can identify question-relevant evidence and form appropriate semantic judgments. OpenEQA covers open-ended scene questions; UniEQA spans multiple embodied questions and skills; and RoboVQA evaluates action, state, outcome, and task understanding from videos of robotic and human activity. These benchmarks overlap in scope, so the grouping organizes the analysis without assuming that each benchmark measures an isolated capability.

\paragraph{Temporal and Event Reasoning.}
Cosmos~\cite{nvidia2025cosmosreason} evaluates the understanding of events, actions, and temporal relations in physical scenes. Relevant questions in RoboVQA and VSI offer additional evidence of temporal reasoning. Each benchmark contributes only once to benchmark-level aggregates; dimension-level scores, when reported, require a frozen mapping from questions to capabilities.

\paragraph{Long-Horizon Action Organization.}
ET-Plan-Bench~\cite{zhang2024etplan} and the EB-ALFRED and EB-Habitat tasks from EmbodiedBench~\cite{yang2025embodiedbench} assess goal-directed reasoning through complete-plan generation and feedback-conditioned action selection. The ET-Plan-Bench archive used here is marked official-agent, with success inherited from the upstream exploration run; we do not equate this configuration with feedback-free, one-shot complete-plan generation. EB-ALFRED and EB-Habitat provide current observations, action history, and available feedback during interaction and assess subsequent action selection and task completion. These settings probe complementary aspects of action organization; claims about closed-loop behavior rely on the interactive evaluations.

\paragraph{Point Grounding and Affordance.}
VABench-Point (VABench-P)~\cite{yuan2025fsd,ni2025embodiedarena} and Where2Place-point~\cite{yuan2024robopoint} evaluate instruction-conditioned localization of actionable targets or valid placement regions. The model predicts normalized coordinates, and scoring measures their agreement with the official target region. VABench-P defines target regions through bounding-box annotations, whereas Where2Place uses region masks. These scoring annotations are distinct from the point outputs requested from the model.

\subsubsection{Training Data}

Both training stages use the corpora that the self-evolution loop of Section~\ref{sec:engine} converged on, and no model in this section was trained on data revised for it specifically. \textbf{Stage~A} is a joint release of about $462$k physical records over $295$k unique identities, split $90\!:\!10$ by physical records between question answering and planning; it was derived from the decontaminated release of Section~\ref{sec:engine} by removing $681$ quarantined or invalid records and then aligning $47$k VABench-P samples to the frozen eight-point XML evaluation contract. \textbf{Stage~B} is a joint post-training release of about $105$k physical records, each unique identity materialized exactly twice; it rebalances toward planning recovery rather than toward point grounding alone, so task planning takes a quarter of its physical records and close to two thirds of its assistant supervision tokens. Table~\ref{tab:supervision_methods_en} gives the construction method per capability dimension, and Appendix~\ref{app:engine} the realized composition of each stage together with the released record counts.

The corpora are assembled from public train splits and simulator rollouts, organized by the same five capability dimensions used for diagnosis and evaluation. Only train splits enter them, and the sources below are named by what each contributes rather than by the benchmark it shares a name with.

\paragraph{Spatial and relational supervision.} EmbSpatial~\cite{du2024embspatial} contributes multiple-choice spatial relations posed from embodied viewpoints, and VSI~\cite{yang2024vsibench} layout, distance and size questions answered from video of a walked-through scene. Euclid30K~\cite{lian2025euclid} adds geometric problems used as a surrogate task for spatial perception. RefSpatial~\cite{zhou2025roborefer} provides spatial referring expressions with reasoning traces, HANDAL~\cite{guo2023handal} manipulable object categories with pose and affordance annotations, and RoboRefIt~\cite{lu2023vlgrasp} language-conditioned object referring in cluttered indoor scenes.

\paragraph{Scene and state supervision.} RoboVQA~\cite{sermanet2023robovqa} contributes long-horizon question answering over robot and human activity video, and Robo2VLM~\cite{chen2025robo2vlm} question answering derived from in-the-wild robot manipulation datasets. EgoPlan-IT~\cite{chen2023egoplan} adds planning questions over egocentric activity video, RoboFAC~\cite{ye2025robofac} robot failure cases with analysis and correction annotations, and EO-Data~\cite{qu2025eo1} instruction-following data from a unified embodied foundation model. MM-IFInstruct~\cite{ding2025mmifengine} supplies multimodal instruction-following data, and LLaVA instruction data~\cite{liu2023llava} general visual instruction tuning, which keeps competence outside the embodied distributions the other sources cover.

\paragraph{Temporal and event supervision.} Cosmos-Reason~\cite{nvidia2025cosmosreason} supplies physical common-sense and event items over short video, here grounded mainly in RoboVQA clips.

\paragraph{Point grounding and affordance supervision.} FSD~\cite{yuan2025fsd} is the main source of eight-point supervision, contributing manipulation data that bridges reasoning and decision. RoboPoint and Where2Place~\cite{yuan2024robopoint} contribute spatial affordance prediction, namely target points and valid placement regions. HOI4D~\cite{liu2022hoi4d} adds egocentric category-level human-object interaction, and CoSyn~\cite{yang2025cosyn} code-guided synthetic text-rich images.

\paragraph{Long-horizon supervision.} ALFRED~\cite{shridhar2020alfred} contributes household instruction-following trajectories in simulation, and Habitat~\cite{savva2019habitat} simulator rollouts, which are retained only when replay reaches the target state. Public planning corpora and in-house planning annotations complete the partition.

\subsubsection{Metrics and Aggregation}

Multiple-choice tasks use option accuracy. Numerical questions follow their benchmark-specific scoring rules; VSI combines exact matching with Mean Relative Accuracy (MRA) according to question type. Open-ended answers use the LLM-based evaluators in the current evaluation implementation. Pointing tasks measure the fraction of predicted points inside the target region. The planning benchmarks use their respective task-success metrics, with format validity, task progress, and invalid-action rates retained as diagnostics. Table~\ref{tab:metrics_en} summarizes the current metrics.

\begin{table}[!htbp]
\centering
\caption{Primary metrics for each evaluation benchmark.}
\label{tab:metrics_en}
\small
\setlength{\tabcolsep}{4pt}
\fitwidth{%
\begin{tabular}{llll}
\toprule
\textbf{Dimension} & \textbf{Benchmark} & \textbf{Primary Metric} & \textbf{Input Modality} \\
\midrule
\multirow{3}{*}{Spatial \& Relational} & EmbSpatial & Exact-choice accuracy & Image \\
 & VSI & Exact match / MRA & Video / scene sequence \\
 & ERQA & Exact-choice accuracy & Ordered multi-image \\
\midrule
\multirow{3}{*}{Scene \& State} & OpenEQA & Frozen LLM-match & Egocentric observation \\
 & UniEQA & 12-skill LLM-judged macro & Multimodal scene \\
 & RoboVQA & Frozen LLM-match & Robot video \\
\midrule
Temporal \& Event & Cosmos & Exact-choice accuracy & Short video / frame sequence \\
\midrule
\multirow{3}{*}{Long-Horizon Action Org.} & ET-Plan-Bench & Task success & Text+Image \\
 & EB-ALFRED & Success Rate & Interactive stream \\
 & EB-Habitat & Success Rate & Interactive stream \\
\midrule
\multirow{2}{*}{Point \& Affordance} & VABench-P & Target-box hit accuracy & Image \\
 & Where2Place-point & Spatial-reference mask accuracy & Image \\
\bottomrule
\end{tabular}%
}
\end{table}

We distinguish benchmark scores, capability-level analyses, and official leaderboard aggregates. The arithmetic mean over the EQA suite summarizes nine benchmarks and is not interchangeable with an official score aggregated over fine-grained capabilities. The planning-suite mean likewise provides a within-suite summary. The five dimensions organize capability coverage; benchmark grouping alone does not define five independent, equally weighted capability scores. We also report a plain equally weighted mean over all twelve benchmarks as an overall summary, which is not a leaderboard aggregate: the EQA benchmarks are score-like metrics whereas the long-horizon benchmarks are task success rates, so the combined mean should be read together with the two group means rather than in place of them. Because the EQA suite mixes perceptual and pointing benchmarks, the main table additionally reports means by competency: perception over seven benchmarks, planning over three, and execution over the two point-grounding benchmarks. This is the partition the capability analysis of this report is organized around.

The main table reports updated individual results and aggregate values. Every listed model has a score on all twelve benchmarks; a dash marks a measured zero, which contributes zero to the sum and counts toward the averaging denominator. Every mean is therefore an equally weighted average over the same benchmarks with the same denominator across models. This notation is specific to the main results table; a result missing because of evaluation infrastructure failure would not be shown as a dash and is not treated as a measured zero.

\subsubsection{Inference Protocol}

ABot-ER uses one set of model weights across tasks and selects an internal branch through an input condition. Following Section~\ref{sec:architecture}, the condition is the data task family of the record: \texttt{[EQA]} for embodied question answering and \texttt{[TASK\_PLAN]} for task planning. Point-grounding benchmarks are evaluated under the former condition: their supervision is collected as part of the question-answering family, even though the answer is a coordinate rather than text. The token is injected by the loader rather than stored in the records.

The audited point predictions use coordinates normalized to $[0,1]$ relative to the original image. Where2Place uses tuple lists, whereas the current VABench archive uses tagged eight-point lists; all 300 answers repeat one location. These cannot be interpreted as eight independent predictions. Evaluation adapters apply deterministic numerical or format conversions, maintaining consistency between target regions and image preprocessing coordinates. The audited EQA configurations specify greedy autoregressive decoding without multiple-sample ensembling. EB archives contain a reasoning\_and\_reflection field, so an absence of explicit reasoning is not assumed across all evaluations; the provenance of its original online calls remains to be verified, precluding inferences about native closed-loop behavior. A single generation episode still requires token-by-token decoding and does not imply a single network forward pass.

\subsubsection{Baseline Models}

We compare ABot-ER with Cosmos-Reason-1-7B~\cite{nvidia2025cosmosreason}, EmbodiedBrain-7B~\cite{zou2025embodiedbrain}, Cambrian-S-7B~\cite{yang2025cambrians}, IflyBot-VLM-8B~\cite{nie2025iflybot}, Mimo-Embodied-7B~\cite{hao2025mimo}, RynnBrain-8B~\cite{rynn2026brain}, Hy-Embodied-RxBrain-1.0~\cite{liang2026rxbrain}, RynnBrain1.1-9B~\cite{li2026rynnbrain11}, and Cosmos3-Nano~\cite{nvidia2026cosmos3} to assess its relative performance across embodied capabilities. We report available benchmark results for each model without treating missing entries as zero scores.

Public-baseline scores are drawn from the Embodied Arena~\cite{ni2025embodiedarena} snapshot of August 12, 2026, supplemented with results available in the local snapshot of September 8, 2026. Our inference settings apply only to results explicitly reevaluated under this protocol. Public entries may differ in parameter count, training data, and original evaluation configuration; these comparisons assess performance, whereas mechanistic attribution requires controlled experiments.

\subsection{Results Across Embodied Capabilities}

The central result in Table~\ref{tab:main_results_en} is that competitive EQA and long-horizon performance coexist within a single ABot-ER model. Read by competency, ABot-ER leads all three means: perception by $3.94$ points over the strongest baseline, planning by $4.20$, and execution by $11.81$. Those three strongest baseline values belong to three different models: RynnBrain1.1-9B on perception, Cosmos3-Nano on planning, and Mimo-Embodied-7B on execution. No single competitor is therefore the one to beat. No baseline matches ABot-ER on both perception and planning at once (Figure~\ref{fig:teaser}). Relative to the strongest listed public baseline with complete results for each suite, ABot-ER improves the EQA and planning means by 9.01 and 4.20 points, respectively, and achieves the best listed result on six of twelve benchmarks. This breadth extends beyond an isolated task-specific advantage. Aggregate comparisons are restricted to identical benchmark sets and do not equate available-entry means or suite averages with official leaderboard scores.

\begin{table}[!t]
\centering
\renewcommand{\arraystretch}{1.05}
\caption{Main results across five capability dimensions and twelve benchmarks. Bold marks the best listed score; \texttt{--} denotes a valid zero and is included in the means. All means are equally weighted, and every listed model has valid results on all twelve benchmarks, so the denominators are identical across models. Perception (7) and Execution (2) together make up EQA (9). All (12) mixes score-like and success-rate metrics and should be read together with the rows above it.}
\label{tab:main_results_en}
\scriptsize
\setlength{\tabcolsep}{2.8pt}
\fitwidth{%
\begin{tabular}{l|c|ccccccccc}
\toprule
\textbf{Benchmark} & \makecell[b]{\textbf{ABot-ER}\\9B} & \makecell[b]{Cosmos\\Reason\\7B} & \makecell[b]{Embodied\\Brain\\7B} & \makecell[b]{Cambrian\\S\\7B} & \makecell[b]{IflyBot\\VLM\\8B} & \makecell[b]{Mimo\\Embodied\\7B} & \makecell[b]{Rynn\\Brain\\8B} & \makecell[b]{Hy-Emb.\\RxBrain\\1.0} & \makecell[b]{Rynn\\Brain 1.1\\9B} & \makecell[b]{Cosmos3\\Nano} \\
\midrule
\multicolumn{11}{l}{\cellcolor{gray!6}\textit{Spatial \& Relational Reasoning}} \\
\addlinespace[1pt]
EmbSpatial & 81.68 & 66.96 & 74.37 & 69.13 & 75.05 & 77.59 & 81.46 & 80.38 & \textbf{82.53} & 82.03 \\
VSI & 63.30 & 27.41 & 37.05 & 63.39 & 30.47 & 43.86 & 67.55 & 0.16 & \textbf{71.59} & 48.65 \\
ERQA & \textbf{47.31} & 40.28 & 38.27 & 35.23 & 38.38 & 39.50 & 40.40 & 38.58 & 44.02 & 46.09 \\
\addlinespace[5pt]
\multicolumn{11}{l}{\cellcolor{gray!6}\textit{Scene \& State Understanding}} \\
\addlinespace[1pt]
OpenEQA & 63.33 & 35.17 & 51.79 & 53.19 & 52.95 & \textbf{65.85} & 57.06 & 40.90 & 61.60 & 49.22 \\
UniEQA & 50.66 & 38.38 & 39.02 & 43.50 & 40.63 & 43.93 & 49.68 & 39.51 & \textbf{56.62} & 49.57 \\
RoboVQA & \textbf{71.79} & 53.75 & 63.19 & 37.77 & 53.60 & 66.09 & 64.31 & 19.66 & 50.63 & 56.08 \\
\addlinespace[5pt]
\multicolumn{11}{l}{\cellcolor{gray!6}\textit{Temporal \& Event Reasoning}} \\
\addlinespace[1pt]
Cosmos & \textbf{68.96} & 51.13 & 46.68 & 44.61 & 54.95 & 50.27 & 54.84 & 35.89 & 52.42 & 63.18 \\
\addlinespace[5pt]
\multicolumn{11}{l}{\cellcolor{gray!6}\textit{Task Planning \& Execution}} \\
\addlinespace[1pt]
ET-Plan & 61.15 & 1.32 & 65.89 & 39.35 & 30.75 & 68.17 & 66.39 & 20.02 & 56.35 & \textbf{69.55} \\
EB-ALFRED & \textbf{26.67} & 2.00 & 0.33 & -- & 0.33 & 17.33 & 0.33 & -- & 5.00 & 16.00 \\
EB-Habitat & \textbf{37.00} & 17.00 & 8.67 & 3.00 & 4.67 & 21.00 & 13.00 & -- & 10.33 & 26.67 \\
\addlinespace[5pt]
\multicolumn{11}{l}{\cellcolor{gray!6}\textit{Point Grounding \& Affordance}} \\
\addlinespace[1pt]
VABench-P & \textbf{63.33} & 4.33 & 2.58 & 0.50 & 11.08 & 46.90 & 9.00 & 3.58 & 8.38 & 47.54 \\
W2P & 70.79 & 5.21 & 34.45 & 10.94 & 70.76 & 63.60 & \textbf{75.48} & 40.50 & 70.20 & 57.68 \\
\midrule
\multicolumn{11}{l}{\cellcolor{gray!6}\textit{Means}} \\
\addlinespace[1pt]
Perception (7) & \textbf{63.86} & 44.73 & 50.05 & 49.55 & 49.43 & 55.30 & 59.33 & 36.44 & 59.92 & 56.40 \\
Planning (3) & \textbf{41.61} & 6.77 & 24.96 & 14.12 & 11.92 & 35.50 & 26.57 & 6.67 & 23.89 & 37.41 \\
Execution (2) & \textbf{67.06} & 4.77 & 18.52 & 5.72 & 40.92 & 55.25 & 42.24 & 22.04 & 39.29 & 52.61 \\
\addlinespace[2pt]
EQA (9) & \textbf{64.57} & 35.85 & 43.04 & 39.81 & 47.54 & 55.29 & 55.53 & 33.24 & 55.33 & 55.56 \\
All (12) & \textbf{58.83} & 28.58 & 38.52 & 33.38 & 38.63 & 50.34 & 48.29 & 26.60 & 47.47 & 51.02 \\
\bottomrule
\end{tabular}%
}
\end{table}

Within perceptual understanding, ABot-ER shows particular strength in interpreting embodied scenes and their temporal dynamics. It improves over the strongest listed baselines by 5.70 points on RoboVQA and 5.78 points on Cosmos, while also achieving the best listed ERQA result. These complementary gains extend beyond answering scene-related questions to reasoning about events and their temporal relationships. This pattern characterizes a strength in action-relevant embodied understanding.

The long-horizon results show that this breadth is not confined to question answering. ABot-ER achieves the highest planning-suite mean among listed models with complete results, including the best listed results on both EB-ALFRED and EB-Habitat, with a 9.34-percentage-point margin on EB-ALFRED over the strongest listed baseline. Competitive perceptual and long-horizon performance thus coexist within the same model, without requiring leadership on every individual benchmark. This conclusion concerns the reported task outcomes, not stronger feedback correction or native closed-loop execution.

Point grounding extends this coverage to spatial predictions relevant to manipulation. ABot-ER exceeds the strongest listed baseline, Cosmos3-Nano, by 15.79 percentage points on VABench-P, showing that its strengths extend beyond semantic answers to instruction-conditioned target localization. Its relative standing on Where2Place is less favorable, so this gain does not imply leadership across all grounding tasks. Taken together, the main results support broad coverage of perception, action organization, and spatial grounding within one model; point-hit rates remain distinct from control precision and physical manipulation success.

\FloatBarrier
\subsection{One Recipe Across Model Sizes}
\label{sec:scaling}

The self-evolution loop of Section~\ref{sec:engine} was run once, on the 9B backbone, and its product is a pair of corpora: the Stage~A joint release and the Stage~B targeted release. This subsection asks whether that product transfers. We train four further backbones, namely dense models of 2B, 4B, and 27B parameters and a 35B mixture-of-experts model with 3B active parameters, with exactly one Stage~A pass and one Stage~B pass each, on the same two releases, with no per-size diagnosis, corpus revision, or mixture adjustment. Every model uses the task-conditioned architecture of Section~\ref{sec:architecture} and is selected by the guardrailed selection rule of Section~\ref{sec:training}. Table~\ref{tab:scaling_config_en} lists the configurations and Table~\ref{tab:scaling_results_en} the results on the twelve benchmarks.

\begin{table}[!htbp]
\centering
\caption{Backbones trained with the one-pass recipe: one Stage~A run, then one joint post-training run from the selected Stage~A checkpoint. Every model uses the same two releases and the same task-conditioned architecture.}
\label{tab:scaling_config_en}
\small
\renewcommand{\arraystretch}{1.1}
\setlength{\tabcolsep}{4pt}
\fitwidth{%
\begin{tabular}{llcc}
\toprule
\textbf{Backbone} & \textbf{Params} & \textbf{Residual layers} & \textbf{GPUs} \\
\midrule
Qwen3.5-2B & 2B & 11 / 15 / 19 / 23 & 16 \\
Qwen3.5-4B & 4B & 19 / 23 / 27 / 31 & 16 \\
Qwen3.5-9B & 9B & 19 / 23 / 27 / 31 & 32 \\
Qwen3.8-27B & 27B & 51 / 55 / 59 / 63 & 32 \\
Qwen3.6-35B-A3B & 35B (3B active) & 27 / 31 / 35 / 39 & 32 \\
\bottomrule
\end{tabular}%
}
\par\smallskip
\parbox{\linewidth}{\footnotesize
Every reported checkpoint is chosen by the guardrailed selection rule of Section~\ref{sec:training}; Stage~A checkpoints are selected jointly across capabilities and are not the peak of either suite taken alone. Residual layer indices are zero-based and are in every case the last four full-attention blocks, whose depths are 24, 32, 32, 64 and 40. Shared across backbones: Stage~A learning rate $4.0\times10^{-6}$, Stage~B learning rate $3.0\times10^{-7}$, global batch 128, residual rank 16, frozen vision encoder with the connector, decoder and task-conditioned residuals trainable, and optimizer and scheduler state reset at the stage boundary.}
\end{table}

\begin{table}[!htbp]
\centering
\caption{Twelve-benchmark results of the one-pass recipe across backbone sizes, each at its best checkpoint. Bold marks the best score in each row.}
\label{tab:scaling_results_en}
\scriptsize
\renewcommand{\arraystretch}{1.0}
\setlength{\tabcolsep}{4pt}
\begin{tabular*}{0.9\linewidth}{@{\extracolsep{\fill}}lrrrrr@{}}
\toprule
\textbf{Benchmark} & \textbf{2B} & \textbf{4B} & \textbf{9B} & \textbf{27B} & \textbf{35B-A3B} \\
\midrule
\multicolumn{6}{l}{\cellcolor{gray!6}\textit{Spatial \& Relational Reasoning}} \\
EmbSpatial & 78.87 & 79.53 & 81.68 & 82.39 & \textbf{82.69} \\
VSI & 49.85 & 58.85 & 63.30 & 65.62 & \textbf{66.85} \\
ERQA & 36.93 & 40.36 & 47.31 & \textbf{49.65} & 47.84 \\
\addlinespace[2pt]
\multicolumn{6}{l}{\cellcolor{gray!6}\textit{Scene \& State Understanding}} \\
OpenEQA & 56.21 & 60.10 & 63.33 & \textbf{67.25} & 65.33 \\
UniEQA & 44.43 & 48.50 & 50.66 & \textbf{53.45} & 52.60 \\
RoboVQA & 68.64 & 72.32 & 71.79 & 73.39 & \textbf{74.22} \\
\addlinespace[2pt]
\multicolumn{6}{l}{\cellcolor{gray!6}\textit{Temporal \& Event Reasoning}} \\
Cosmos & 55.58 & 68.69 & \textbf{68.96} & 67.71 & 65.73 \\
\addlinespace[2pt]
\multicolumn{6}{l}{\cellcolor{gray!6}\textit{Task Planning \& Execution}} \\
ET-Plan & 64.41 & 60.75 & 61.15 & \textbf{69.11} & 67.77 \\
EB-ALFRED & 13.67 & 30.00 & 26.67 & \textbf{37.00} & 31.00 \\
EB-Habitat & 24.33 & 34.33 & 37.00 & \textbf{48.67} & 40.67 \\
\addlinespace[2pt]
\multicolumn{6}{l}{\cellcolor{gray!6}\textit{Point Grounding \& Affordance}} \\
VABench-P & 54.00 & 61.33 & \textbf{63.33} & 62.67 & 62.33 \\
Where2Place & 52.26 & 66.47 & 70.79 & \textbf{73.90} & 70.91 \\
\midrule
EQA (9) & 55.20 & 61.80 & 64.57 & \textbf{66.23} & 65.39 \\
Planning (3) & 34.14 & 41.69 & 41.61 & \textbf{51.59} & 46.48 \\
All (12) & 49.93 & 56.77 & 58.83 & \textbf{62.57} & 60.66 \\
\addlinespace[2pt]
Perception (7) & 55.79 & 61.19 & 63.86 & \textbf{65.64} & 65.04 \\
Execution (2) & 53.13 & 63.90 & 67.06 & \textbf{68.29} & 66.62 \\
\bottomrule
\end{tabular*}
\par\smallskip
\parbox{\linewidth}{\footnotesize
One Stage~A and one Stage~B pass on the same two releases for every model. All~(12) is the plain twelve-benchmark mean, not the average of EQA~(9) and Planning~(3). Perception~(7), Execution~(2) and All~(12) are derived from the per-benchmark scores.}
\end{table}

Two readings of Table~\ref{tab:scaling_results_en} matter. The first is whether the profile transfers, and it does: every backbone trained on the converged releases, with no size-specific diagnosis or revision, lands with all five dimensions populated rather than with the lopsided profile the pretrained models start from. The 2B model already reaches usable levels on all five, and each larger backbone keeps that shape while lifting it.

The second is how the profile moves with scale, where the three competencies do not move together. From 2B to 27B the perception mean rises by $9.85$ points, the execution mean by $15.16$, and the planning mean by $17.45$: perception is closest to its ceiling at small size, while long-horizon action organization is what parameters buy most. EmbSpatial illustrates the first, gaining $3.52$ points across a more than tenfold increase in parameters, and EB-Habitat the second, gaining $24.34$. The mixture-of-experts entry shows that the recipe is not tied to a dense backbone: with 3B active parameters it sits between the 9B and 27B dense models on the twelve-benchmark mean and on both suite means.

Taken together with the one-pass comparison of Section~\ref{sec:training}, this study is the second half of the same argument: the loop's product is a corpus, the corpus is what a new model needs, and the diagnosis that produced it does not have to be repeated for each backbone.

\subsection{Effect of the Task-Conditioned Adaptive Training Strategy}

This subsection reports what the second stage of the released recipe contributes, and what task conditioning adds on a backbone that already routes. Table~\ref{tab:task_conditioned_training_en} isolates the stage: it compares the Stage~A checkpoint the released model starts from against the released checkpoint, both on the task-conditioned open-source line.

\begin{table}[!htbp]
\centering
\caption{What the joint post-training stage adds. The first column is the Stage~A checkpoint the released model starts from, the second is the released checkpoint, and $\Delta$ is the difference. Both are the task-conditioned open-source line; no unconditioned run enters this comparison. Bold marks the higher of the two.}
\label{tab:task_conditioned_training_en}
\begingroup
\small
\renewcommand{\arraystretch}{1.12}
\setlength{\tabcolsep}{6pt}
\begin{tabular*}{0.8\linewidth}{@{\extracolsep{\fill}}lrrr@{}}
\toprule
\textbf{Benchmark} & \makecell[r]{\textbf{Stage A}\\selected} & \textbf{Released} & $\Delta$ \\
\midrule
EmbSpatial & \textbf{81.70} & 81.68 & $-0.02$ \\
VSI & 63.11 & \textbf{63.30} & $+0.19$ \\
ERQA & 45.39 & \textbf{47.31} & $+1.92$ \\
OpenEQA & 63.14 & \textbf{63.33} & $+0.19$ \\
UniEQA & 50.63 & \textbf{50.66} & $+0.03$ \\
RoboVQA & \textbf{72.29} & 71.79 & $-0.50$ \\
Cosmos & 67.58 & \textbf{68.96} & $+1.38$ \\
VABench-P & 61.67 & \textbf{63.33} & $+1.66$ \\
Where2Place & \textbf{72.90} & 70.79 & $-2.11$ \\
\midrule
ET-Plan-Bench & 60.95 & \textbf{61.15} & $+0.20$ \\
EB-ALFRED & 23.67 & \textbf{26.67} & $+3.00$ \\
EB-Habitat & 33.67 & \textbf{37.00} & $+3.33$ \\
\midrule
\textbf{EQA (9)} & 64.27 & \textbf{64.57} & $+0.30$ \\
\textbf{Planning (3)} & 39.43 & \textbf{41.61} & $+2.18$ \\
\bottomrule
\end{tabular*}
\endgroup
\end{table}

The effect has the shape the two-stage design of Section~\ref{sec:training} predicts and the composition of the post-training release implies. That release devotes a quarter of its physical records and close to two thirds of its assistant supervision tokens to planning, and the gains land there: the two interactive planning benchmarks rise by $3.00$ and $3.33$ points and the planning mean by $2.18$, against $0.30$ on the nine-benchmark question-answering mean. The perception benchmarks, already close to saturation at the end of Stage~A, move by at most $1.92$ in either direction. The stage is therefore a redistribution toward the diagnosed deficit with the saturated dimensions held roughly level, not a uniform improvement, and the $2.11$-point regression it leaves on Where2Place is the price of that concentration.

\paragraph{Task conditioning on the released mixture-of-experts backbone.} A mixture-of-experts backbone already routes computation at the token level, so whether explicit sequence-level task conditioning still contributes on top of it is a fair question, and on this backbone it is a question about a released model: the 35B-A3B entry of Table~\ref{tab:scaling_results_en} is trained with the same task-conditioned recipe and activates 3B parameters per token, yet reaches a higher twelve-benchmark mean than the 9B dense model, a higher planning mean than any backbone below 27B, and one more leading result than the 9B model. We ran the ablation on this backbone twice, under corpora at two different stages of repair, and the two runs answer different halves of the question.

\begin{table}[!htbp]
\centering
\caption{Task-conditioning ablation on Qwen3.6-35B-A3B under two corpora. Each pair differs only in whether the route tokens and conditional residuals are enabled, with data, optimization, and evaluation held fixed. The first pair is a single Stage~A comparison at a matched training step on a corpus whose point supervision still carried unaligned coordinate conventions, which is why its VABench-P scores sit near $20$. The second pair is measured after the protocol unification of Section~\ref{sec:protocol}, over nine matched pre-training and eight matched post-training checkpoints, each post-training run continuing from its own pre-training checkpoint at the same step.}
\label{tab:moe_ablation_en}
\small
\renewcommand{\arraystretch}{1.15}
\setlength{\tabcolsep}{5pt}
\fitwidth{%
\begin{tabular}{lrrr}
\toprule
 & \textbf{Plain} & \textbf{Task-cond.} & $\Delta$ \\
\midrule
\multicolumn{4}{l}{\cellcolor{gray!6}\textit{Corpus with unaligned output conventions, matched Stage~A step}} \\
\addlinespace[2pt]
EQA (9) & 58.62 & \textbf{60.24} & $+1.62$ \\
Planning (3) & 31.89 & \textbf{32.50} & $+0.61$ \\
All (12) & 51.94 & \textbf{53.30} & $+1.36$ \\
Benchmarks improved & \multicolumn{3}{r}{11 of 12} \\
\addlinespace[5pt]
\multicolumn{4}{l}{\cellcolor{gray!6}\textit{Corpus after protocol unification, 17 matched checkpoints}} \\
\addlinespace[2pt]
EQA (9), best checkpoint of each run & \textbf{66.94} & 66.27 & $-0.67$ \\
Planning (3), best checkpoint of each run & \textbf{42.87} & 41.94 & $-0.93$ \\
ET-Plan-Bench & \multicolumn{3}{r}{higher with conditioning at 17 of 17, mean $+3.3$} \\
\bottomrule
\end{tabular}%
}
\end{table}

On the corpus that still carried unaligned output conventions, conditioning improved eleven of the twelve benchmarks and raised the twelve-benchmark mean by $1.36$. On the repaired corpus that gain is no longer there: whether the two runs are compared at matched steps or each at its own best checkpoint, the suite means come out slightly in favour of the unconditioned run. What survives the repair is narrow and completely consistent: ET-Plan-Bench, the one benchmark that asks only for a complete plan, is higher with conditioning at every one of the seventeen matched checkpoints, by $3.3$ points on average, which is the output regime the \texttt{[TASK\_PLAN]} route conditions.

The practical reading is that task conditioning pays when the training corpus is not yet internally consistent, such as when supervision for the same quantity still carries conflicting conventions, and that its value as a source of aggregate gains declines as that inconsistency is removed. On that reading the conditional residuals and the protocol evolution of Section~\ref{sec:protocol} act on the same conflict from different sides, one in the parameters and one in the data. The two corpora differ in more than one respect, so we state this as the reading we act on rather than a controlled result; what it recommends is concrete enough to act on, namely to condition while a corpus is still being repaired, and to expect the return to shrink once it is.

\FloatBarrier
\subsection{Analysis of Data-Model Self-Evolution}

The central challenge of self-evolution is to expand capabilities while retaining existing strengths, rather than to increase every benchmark score at every round. Consistent with our capability-oriented evaluation, we first examine the overall capability profile and then use benchmark trajectories and within-round checkpoints to investigate the changes underlying it.

The versions compared in this subsection were not produced under a common training budget. Table~\ref{tab:round_configs_en} records what each one was, so that a difference between versions is not read as a difference in data strategy alone: the rounds vary by an order of magnitude in training budget and learning rate, and most of them restart from the pretrained model rather than continuing the previous version. The comparisons below therefore describe how the released line of models developed, not a controlled sweep over data recipes.

\begin{table}[!htbp]
\centering
\caption{Training configuration of the versions compared in this subsection, named as in Figures~\ref{fig:capability_evolution_summary_en} and~\ref{fig:evolution_trajectories_en}. R3 is omitted: its planning score is affected by the contamination described in Section~\ref{sec:case_c}. Records are physical training records of the round's release.}
\label{tab:round_configs_en}
\small
\renewcommand{\arraystretch}{1.12}
\setlength{\tabcolsep}{4pt}
\fitwidth{%
\begin{tabular}{lrrll}
\toprule
\textbf{Round} & \textbf{Records} & \textbf{GPUs} & \textbf{LR} & \textbf{Weight origin} \\
\midrule
Base & --- & --- & --- & Pretrained model \\
R1 & 70,526 & 80 & $8.33\times10^{-6}$ & Pretrained model \\
R2 & 288,028 & 128 & $1.33\times10^{-5}$ & Pretrained model \\
R4 & 463,071 & 64 & $4.0\times10^{-6}$ & Pretrained model \\
R5 & 153,600 & 64 & $8.0\times10^{-7}$ & Continues R4 \\
R6 & 320,000 & 32 & $1.0\times10^{-6}$ & Restarts from R4 \\
\midrule
ABot-ER Stage A & 462,390 & 32 & $4.0\times10^{-6}$ & Pretrained model \\
ABot-ER Stage B & 104,960 & 32 & $3.0\times10^{-7}$ & Continues Stage A \\
\bottomrule
\end{tabular}%
}
\par\smallskip
\parbox{\linewidth}{\footnotesize
The checkpoint carried forward is selected by the guardrailed selection rule of Section~\ref{sec:training} and is generally earlier. R6 is not the released model: the released model is the two-stage run in the lower block, which restarts from the pretrained model on the decontaminated release rather than continuing R6.}
\end{table}

\begin{figure}[!htbp]
\centering
\includegraphics[width=\linewidth]{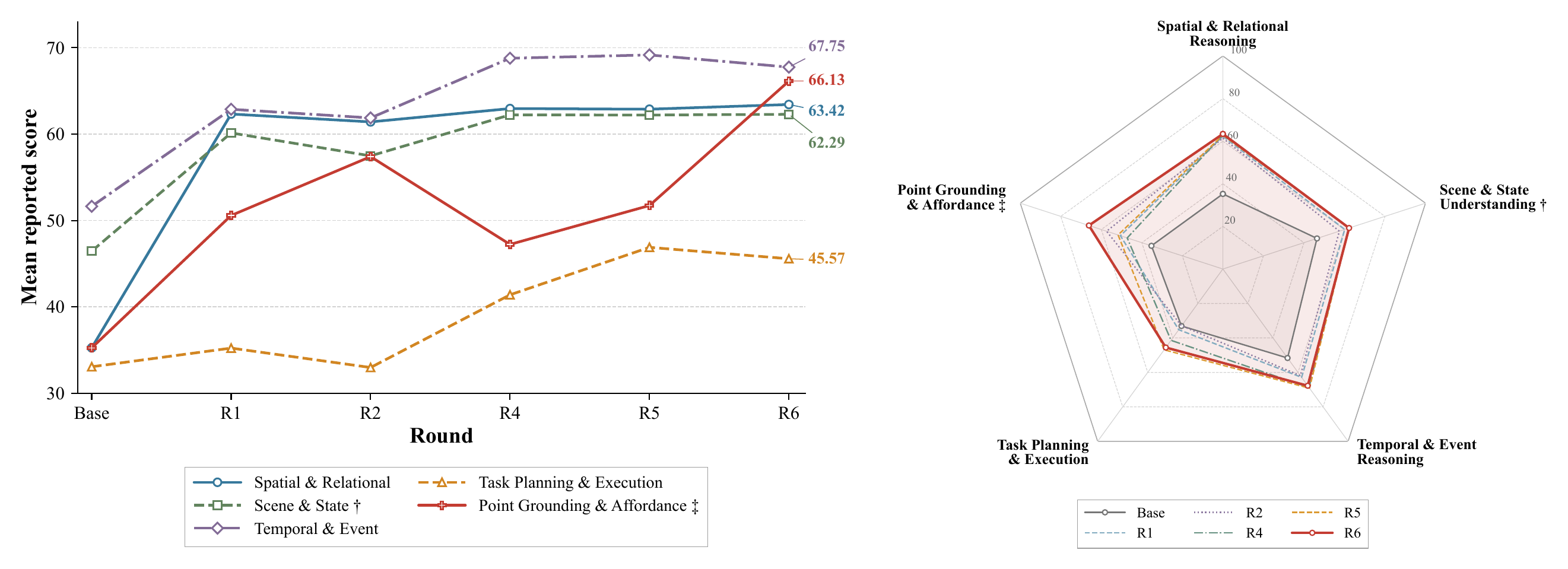}
\caption{Evolution of reported scores across five capability dimensions. Curves show cross-version changes; radar plots show multidimensional profiles. Dimension scores equally average fixed benchmark groups. The rounds are those listed in Table~\ref{tab:round_configs_en}; R6 is not the final model. $\dagger$ retains UniEQA judge-error zeros; $\ddagger$ includes historical VABench-P protocol changes. The figure describes trends, not directly comparable absolute abilities or a radar-area total score.}
\label{fig:capability_evolution_summary_en}
\end{figure}

Figure~\ref{fig:capability_evolution_summary_en} provides a five-dimensional overview of reported scores across selected versions. The curves show how each dimension changes, while the radar plot displays the multidimensional profile of each version. Reported spatial and relational scores stabilize after an early increase, whereas planning and execution scores continue to change more visibly in later stages, revealing different patterns of development across dimensions. These summaries retain historical scoring and protocol effects and therefore serve as an overview, not as evidence that an expanded radar profile represents uniformly validated capability gains.

\begin{figure}[!htbp]
\centering
\includegraphics[width=0.92\linewidth]{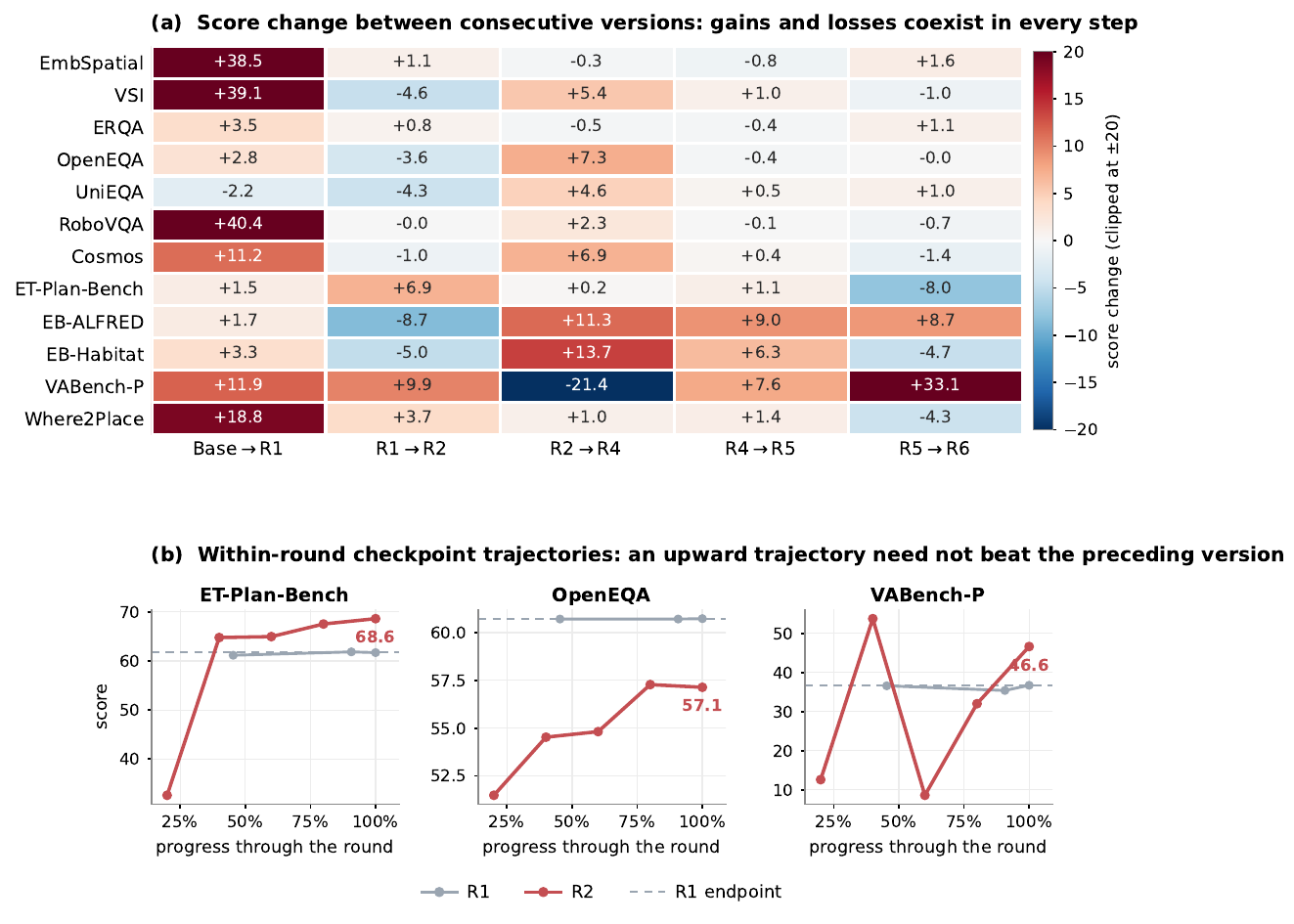}
\caption{Evolution across twelve benchmarks at two scales. (a)~Score change at each transition between the versions of Figure~\ref{fig:capability_evolution_summary_en}; colour is clipped at $\pm 20$ so that the transitions after the first remain legible, and the printed values are unclipped. (b)~Checkpoint trajectories inside R1 and R2 for the three trajectory shapes that occur, with the R1 endpoint as reference. The outcome for all twelve benchmarks is the R1$\to$R2 column of (a), where five of twelve end above the preceding version. All markers are measured results; VABench-P observations in (b) predate the frozen point contract.}
\label{fig:evolution_trajectories_en}
\end{figure}

Capability summaries reveal the overall pattern, but tasks within the same dimension need not improve together. To examine the sources of these changes, Figure~\ref{fig:evolution_trajectories_en} expands the analysis to twelve benchmarks at two scales: panel~(a) records the score change of every benchmark at every transition between the selected versions, and panel~(b) follows the checkpoints inside a single round, distinguishing cross-version gains from within-round recovery.

The key finding of panel~(a) is that broad improvement is not synchronous improvement: every column mixes gains and losses. The selected versions make progress in spatial understanding and robot-centric question answering, yet their task-specific peaks occur at different points. From R5 to R6, for example, EB-ALFRED rises by $8.7$ points while ET-Plan-Bench, Where2Place, and EB-Habitat decline by $8.0$, $4.3$, and $4.7$. Self-evolution is therefore not a single path of improvement on every metric, but a sequence of tradeoffs between capability expansion and retention. The relevant objective is broad competitiveness within the same model, not a composite of benchmark peaks drawn from different versions.

Panel~(b) distinguishes within-round improvement from gains over a preceding version, and the twelve benchmarks fall into three shapes. During R2, ET-Plan-Bench falls to $32.6$ early in the round and recovers to $68.6$, surpassing its R1 endpoint; OpenEQA rises monotonically to $57.1$ without reaching its R1 level of $60.7$; and VABench-P swings between $8.6$ and $53.8$ across consecutive checkpoints, the instability traced in Section~\ref{sec:protocol} to conflicting coordinate supervision. Seven of the twelve benchmarks follow the OpenEQA shape. An upward trajectory within a round is therefore insufficient to establish a better version. Combining both scales avoids measuring gains exclusively from an early low point; where historical scoring anomalies are retained, changes in reported scores must also be distinguished from changes in underlying capability.

A round is therefore judged by the breadth of improvement together with retention rather than by a peak on any single benchmark: the guardrailed selection rule of Section~\ref{sec:training} accepts a checkpoint only when it adds leading results on the targeted benchmarks without breaching a retention guardrail, which is why local regressions appear in the ledger alongside the gains that paid for them.

\FloatBarrier
\subsection{Real-World Deployment}
\label{sec:realrobot}

We further examine how ABot-ER's embodied perception supports physical interaction on a Galaxea R1 Pro, a wheeled dual-arm robot. The platform has 26 degrees of freedom, including two 7-DoF arms with parallel grippers, a 4-DoF torso, and a 6-DoF omnidirectional chassis; it stands approximately 1.7\,m tall and supports a rated payload of 3.5\,kg per arm at a reach of 0.6\,m.\footnote{Galaxea R1 Pro official specifications: \url{https://galaxea-dynamics.com/zh/pages/r1-pro-product-specification}.} ABot-ER serves as the visual question answering (VQA) component of \textbf{ABot-AgentOS}~\cite{tian2026abotagentosgeneralroboticagent}, our robot agent runtime, deployed through vLLM on an external host with a single NVIDIA GeForce RTX 5090D and taking $1920\times1536$ images from the robot's head camera. ABot-AgentOS orchestrates the interaction and invokes \textbf{ABot-M}~\cite{yang2026abotm0vlafoundationmodel,chen2026abotm05unifiedmobilityandmanipulationworld}, a separate vision-language-action (VLA) model, for physical manipulation.

\begin{figure}[!tb]
\centering
\includegraphics[width=0.86\linewidth]{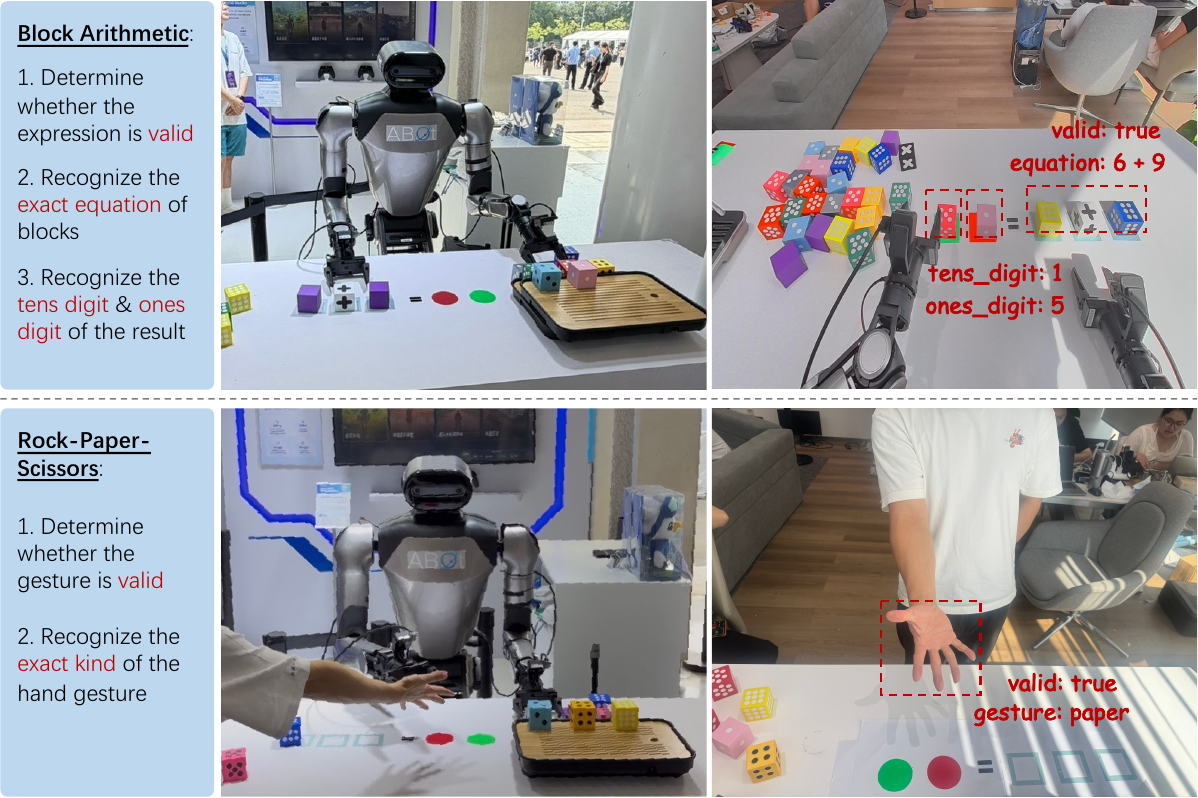}
\caption{\textbf{Real-world interaction with ABot-ER on the Galaxea R1 Pro.} Left: task requirements; middle: physical demonstrations; right: head-camera views annotated with visual recognition results. Top: Block Arithmetic requires expression validation, dot-based number recognition, and result-digit readout, illustrated by $6+9$ and the answer blocks representing $15$. Bottom: Rock-Paper-Scissors requires gesture validity assessment and classification, illustrated by a valid paper gesture. ABot-ER supplies visual state information to ABot-AgentOS, which coordinates decisions and downstream actions.}
\label{fig:realrobot_en}
\end{figure}

\paragraph{The query loop.} ABot-AgentOS queries ABot-ER once before each step and once after, and the two queries ask for different things.

Before the arm moves it asks for two. The first is a reading of the scene: counting the dots on each block, identifying the operator and judging whether the expression is well formed is \emph{Scene and State Understanding}, while recovering the intended operand order is \emph{Spatial and Relational Reasoning}, since the expression faces the participant and therefore reads right to left in the robot's camera image. The second is where to act: a grasp point on the target block, and a placement point inside a free region to the right of the tens digit large enough to receive the ones digit. These are \emph{Point Grounding and Affordance}, returned as normalized $[0,1]$ coordinates under the protocol of Section~\ref{sec:protocol} and consumed directly by ABot-M.

After the arm moves, a second reading reports what the tabletop now holds and whether the step did what it was meant to, which is where \emph{Temporal and Event Reasoning} enters: the assessment separates a block that has been placed from one that has not. ABot-AgentOS decides from it whether to continue or to revise, and a revision re-orders the remaining steps against the current tabletop state, which is \emph{Long-Horizon Action Organization}. The model supplies the judgement; ABot-AgentOS takes the decision.

\paragraph{Demonstrations.} In \textbf{Block Arithmetic}, a participant across the table arranges an arithmetic expression using blocks whose numerical values are encoded by dots rather than printed digits. Figure~\ref{fig:realrobot_en} (top) shows the expression $6+9$ and the visual readout of the result blocks as tens digit~1 and ones digit~5.

In \textbf{Rock-Paper-Scissors}, ABot-ER identifies the participant's hand gesture and judges whether it is a valid game gesture, a reading of the interaction state rather than a category label alone, and therefore again Scene and State Understanding. ABot-AgentOS combines this with the robot's own gesture to determine a win, loss, or draw. Figure~\ref{fig:realrobot_en} (bottom) shows recognition of a valid \emph{paper} gesture.

These qualitative demonstrations connect the capability-oriented training and benchmark evaluation of this report to practical robot use: fine-grained visual evidence and spatial relations supply the state information on which subsequent decisions and actions depend.

\FloatBarrier

%% file: sections/conclusion.tex
\section{Conclusion and Future Work}
\label{sec:conclusion}

This report presented ABot-ER, a unified embodied reasoning model that resolves, from one set of weights, the questions a robot faces between seeing and acting: where things are, what state the scene is in, what just happened, what to do next, and exactly where to act. It is a decoder-only vision-language architecture with a single normalized coordinate protocol, per-benchmark output contracts frozen into the training labels, and a task conditioning mechanism, so that one body of embodied knowledge serves perception, planning, and execution without switching weights or adding a decoding stage at inference. Two obstacles stand between that design and a model that realizes it. Against \textbf{Data-Model Drift}, the Agentic Data-Model Self-Evolution Engine turns corpus construction into a closed loop driven by evaluation feedback, and its central move is to diagnose which aspect of the training data is at fault rather than to answer every weakness with more samples: one shared diagnosis coordinates three revisions that cannot substitute for one another, supplying supervision that is missing, repairing supervision that is contradictory, and reallocating budget to supervision that is under-exposed, with a coordinating agent proposing each diagnosis and deterministic programs deciding what enters training. Against \textbf{Modality Entanglement}, the Task-Conditioned Adaptive Training Strategy overlays lightweight low-rank residuals on the shared backbone, routed by the task family of each record, so that heterogeneous supervision can share representations while receiving the task-specific adaptation intended to reduce interference between them.

In a systematic evaluation across twelve benchmarks covering three core competencies and five capability dimensions, ABot-ER leads all three competency means, namely perception, planning, and execution, among the public models compared, demonstrating that breadth and depth can coexist in unified embodied reasoning. The converged corpus carries much of what the loop produces: a single post-training pass on it reaches a similar twelve-benchmark mean to a chain of six rounds at a fifth of the accumulated post-training, and the same one-pass recipe transfers to backbones from 2B to 27B parameters without any size-specific diagnosis. The same recipe on a mixture-of-experts backbone that activates 3B parameters per token surpasses the 9B dense model on the twelve-benchmark mean, with the conditioning's contribution there concentrated on complete-plan generation. Task conditioning pays while a corpus is still being made internally consistent and returns less once it is: on that backbone it improved eleven of twelve benchmarks before the output protocol was unified and neither suite mean after, with a consistent gain on complete-plan generation throughout. On a wheeled dual-arm platform the model serves as the perception and judgement component of an agent pipeline that delegates motion to a separate action model. Interactive planning is the dimension the aggregate mean describes least well: it is where a single pass on the converged corpus recovers the least of the six-round chain, and where targeted post-training contributes the most.

That gap sets the first of three directions we plan to pursue. The most direct is to extend self-evolution from supervised fine-tuning to learning from environment feedback, using the observed outcomes of model-proposed actions executed by downstream policies. The second is to couple the reasoning and control tiers more tightly while keeping them separate, so that grounded coordinates and plans carry more of the information a vision-language-action model needs, and feedback from execution returns to the reasoning tier as evidence. The third follows from the corpus transferring across backbones: the artifact worth maintaining is that corpus together with its evaluation contract and audit trail, not only the checkpoint trained on it. We release the model weights and this corpus, and of the two we expect the corpus to be the more durable. We hope ABot-ER is useful as a reasoning layer for robot systems, and that the account of how its training data was diagnosed, repaired, and audited is as useful as the model itself.

%% file: sections/contribution.tex
\section{Contributions}
\label{sec:contributions}

\setlength{\parskip}{0pt} 
\setlength{\itemsep}{0pt} 
\setlength{\parsep}{0pt}  
\begin{multicols}{2}
\raggedcolumns

\subsubsection*{Core Contributors}
\begin{itemize}
\item Qiuyu Ding
\item Hengda Xu
\item Zedong Chu
\item Changda Xia
\item Jiajun Shen
\end{itemize}

\subsubsection*{Project Lead}
\begin{itemize}
\item Zedong Chu$^{\dagger}$
\end{itemize}

\subsubsection*{Advisor}
\begin{itemize}
\item Mu Xu
\item Wenbin Tang
\end{itemize}

\columnbreak

\subsubsection*{Contributors}
\begin{itemize}
\item Zihao Guan
\item Zhengchen Yuan
\item Jianfang Zhang
\item Di Yang
\item Jiayi Tian
\item Yuting Xu
\item Fei Wang
\item Shiao Liu
\item Wei Zhang
\item Xiaolong Wu
\end{itemize}

\end{multicols}

\subsubsection*{Acknowledgments}
We extend our sincere gratitude to the broader team for their support, particularly Zixiao Tang, Shihui Su, Zhining Gu, Honglin Han, Minqi Gu, Zeqian Ye, Yang Cai, Linbo Zheng, and Jingjing Ma.

\vfill

\noindent $^{\dagger}$ Corresponding author: chuzedong.czd@alibaba-inc.com

%% file: sections/appendix_engine.tex
\section{Self-Evolution Engine: Full Records}
\label{app:engine}

This appendix collects the complete construction quantities, release audits, data compositions, training configurations, and per-round ledger of the engine described in Section~\ref{sec:engine}. Accounting bases and round attributions are stated in each caption and table note.

\subsection{Supervision Construction and Release Audit}

\begin{table}[!htbp]
\centering
\caption{Supervision construction by capability dimension. The construction column describes the methods used across rounds; historical processing volumes named there belong to the rounds that performed them. The count column is the R6 development release, whose rows sum to its $160{,}000$ unique identities, each materialized twice for $320{,}000$ physical records. It is not the released model, whose Stage~A composition is given in Table~\ref{tab:stage_a_components_en}.}
\label{tab:supervision_construction_en}
\scriptsize
\renewcommand{\arraystretch}{1.1}
\setlength{\tabcolsep}{4pt}
\begin{tabular}{@{}p{0.145\textwidth}p{0.72\textwidth}r@{}}
\toprule
\textbf{Dimension} & \textbf{Construction method} & \makecell[r]{\textbf{Unique,}\\\textbf{R6}} \\
\midrule
Spatial \& Relational &
Train-only supervision from public spatial corpora, cleaned, relabelled, and sampled under constraints; multiple choice constrained to a single A--D token; numeric answers passed through the benchmark parser; equal-weight channels over 21 fine-grained capabilities correct natural source imbalance; in the constrained rebuild, $10{,}373$ released records had their capability metadata corrected &
$46{,}104$ \\
\addlinespace[2pt]
Scene \& State &
Existing train question-answer pairs with answer prefixes and wrapper markup stripped; repetition limited by scene and group quotas. In the targeted release, indoor question answering is additionally constructed deterministically from visible objects, target objects, task instructions, visual state descriptions, and verified action plans of replay-passed episodes, with quotas over seven question types &
$45{,}996$ \\
\addlinespace[2pt]
Temporal \& Event &
Existing video question answering; strict multiple choice preserved as A--D; yes/no items mapped to A/B only where the mapping is determinate, with no model-generated distractors; frame variants of the same video and question share a group exposure budget; the targeted release draws five independent clean train domains at roughly equal share. In R6 the dimension resolves into $3{,}037$ multiple-choice items, which include the $3{,}031$ determinately converted binary items, and $4{,}264$ open-text proxies &
$7{,}301$ \\
\addlinespace[2pt]
Point \& Affordance &
Early rounds: target medoid for dataset point sources, endpoint extraction for object trajectories, deletion of non-object trajectories, normalization of free-space points to $[0,1]$ tuples. After Case~A: only labels with an explicit target region are retained. For the eight-point supervision used with the VABench-P contract, where the source provides eight free points and a bounding box, $70\%$ of labels keep the original eight free points with at least four distinct coordinates and $30\%$ use the medoid of valid original points repeated eight times; Where2Place retains its own one-to-three-point protocol. The earlier rebuild rewrote $247{,}153$ trajectory labels into target-region or endpoint form and dropped $9{,}120$ non-object trajectories from the candidate pool &
$20{,}599$ \\
\addlinespace[2pt]
Long-Horizon Action Org. &
Multi-step and long-horizon windows selected from successful train trajectories with action identifiers and names recorded. For Habitat, each symbolic action is instantiated against the simulator state, its PDDL preconditions are verified, the action is executed in the environment, and only trajectories reaching the target terminal state under replay are retained. For ALFRED, trajectories are drawn from a soft-deduplicated clean split marked as replay-passing. The two source pools hold $36{,}000$ and $24{,}807$ decision-window records respectively, of which about a third are feedback-driven recovery windows; R6 samples $15{,}000$ and $12{,}000$ of them, alongside $13{,}000$ general replay-verified planning records &
$40{,}000$ \\
\bottomrule
\end{tabular}
\end{table}

\begin{table}[!htbp]
\centering
\caption{Hard gates and the result recorded for each, on the two released stages. Both stages close as pass with documented residual risk. Where a quantity was not recorded separately, the cell says so rather than reporting a zero.}
\label{tab:audit_en}
\small
\renewcommand{\arraystretch}{1.18}
\setlength{\tabcolsep}{4pt}
\begin{tabular}{p{0.20\linewidth}p{0.44\linewidth}p{0.26\linewidth}}
\toprule
\textbf{Gate} & \textbf{Requirement} & \textbf{Result and round} \\
\midrule
Evaluation contamination & No detected overlap under the specified identifier and hash-matching checks; unindexed media remain documented residual risk & No exact test-sample match on either stage within the checked identifiers and the available hash coverage. The Stage~A finding comes from the joint audit with its decontaminated parent, since the format-alignment step changed no media, source or identity. The $300$ official point-grounding test anchors were isolated earlier, at source admission \\
\addlinespace[2pt]
Source eligibility & No records drawn from sources ruled ineligible & No prohibited-source record remains on either stage. Three sources had been removed in full, quarantining $23{,}565$ input records, when the decontaminated parent release was built \\
\addlinespace[2pt]
Output contract & Every label matches the contract template and passes the schema and pattern checks & Stage~B: $104{,}960$ of $104{,}960$ labels pass, $100\%$. Stage~A: all $46{,}776$ VABench-P labels are valid eight-point XML, but no round-trip rate under one shared runtime parser was recorded across the remaining benchmarks. A byte-level prompt-match rate was not recorded for either stage \\
\addlinespace[2pt]
Coordinate protocol & No box labels, no non-canonical points, no conflicting protocols for the same question & Zero box outputs and zero non-canonical points on both stages \\
\addlinespace[2pt]
Reference integrity & Every record resolves to an existing media file & Zero media-schema errors on Stage~A, and the media check ran and passed on Stage~B; neither stage saved a separate count of missing references \\
\bottomrule
\end{tabular}
\end{table}

\paragraph{Screening coverage.} Two screens are reported as coverage rather than as a gate. An external perceptual-hash index covers $408{,}502$ of the $707{,}758$ distinct media paths in the joint audit of the two released stages, or $57.72\%$, and $99{,}100$ of the $153{,}146$ paths in the Stage~B audit taken alone, or $64.71\%$; no overlap against evaluation media was found inside the covered index. The corresponding figure for Stage~A taken alone was not recorded. Separately, every one of the $418{,}492$ perceptual hashes embedded in the joint release and the $81{,}735$ embedded in Stage~B was checked; that is a different basis and the two coverages are not additive. Entries missing from the index are carried as residual risk. Semantic near-duplicates and anomalous scores are escalated for case-by-case adjudication rather than resolved by threshold, and the resulting candidates are isolated individually.

\subsection{Stage Compositions and Training Configuration}

\begin{table}[!htbp]
\centering
\caption{Realized composition of the Stage~A joint training set by physical records, grouped by capability dimension. This release does not carry a regenerated token-proxy report, so no token-proxy share is stated for it; the $316.01$M figure of the earlier R4 version does not transfer. Long-horizon action organization is the planning partition, comprising $42{,}236$ general planning records, $1{,}598$ Habitat-grounded and $349$ ALFRED-grounded plans.}
\label{tab:stage_a_components_en}
\small
\renewcommand{\arraystretch}{1.1}
\setlength{\tabcolsep}{6pt}
\fitwidth{%
\begin{tabular}{lrr}
\toprule
\textbf{Capability dimension} & \textbf{Physical} & \textbf{Share} \\
\midrule
Spatial \& Relational & 144,902 & 31.34\% \\
Scene \& State & 150,806 & 32.61\% \\
Temporal \& Event & 11,905 & 2.57\% \\
Point \& Affordance & 110,594 & 23.92\% \\
Long-Horizon Action Org. & 44,183 & 9.56\% \\
\midrule
Total & 462,390 & 100\% \\
\bottomrule
\end{tabular}%
}
\end{table}

\begin{table}[!htbp]
\centering
\caption{Realized composition of the Stage~B Final Joint closure set by physical records. Each unique identity is materialized exactly twice. The release rebalances toward planning recovery under a question-answering guardrail, not toward point grounding alone.}
\label{tab:stage_b_components_en}
\small
\renewcommand{\arraystretch}{1.1}
\setlength{\tabcolsep}{6pt}
\fitwidth{%
\begin{tabular}{lrrr}
\toprule
\textbf{Component} & \textbf{Physical} & \textbf{Unique} & \textbf{Share} \\
\midrule
Task planning & 26,240 & 13,120 & 25.00\% \\
VABench-P & 16,794 & 8,397 & 16.00\% \\
OpenEQA & 12,000 & 6,000 & 11.43\% \\
ERQA & 10,496 & 5,248 & 10.00\% \\
Where2Place & 10,000 & 5,000 & 9.53\% \\
RoboVQA & 8,000 & 4,000 & 7.62\% \\
Cosmos & 6,074 & 3,037 & 5.79\% \\
EmbSpatial & 6,000 & 3,000 & 5.72\% \\
UniEQA & 5,356 & 2,678 & 5.10\% \\
VSI & 4,000 & 2,000 & 3.81\% \\
\midrule
Total & 104,960 & 52,480 & 100\% \\
\bottomrule
\end{tabular}%
}
\par\smallskip
\parbox{\linewidth}{\footnotesize
Component names denote the capability each group is constructed to support rather than the identity of its raw sources; the sources themselves are listed in Section~\ref{sec:experiments}. The task-planning partition holds $12{,}000$ replay-verified ALFRED records, $9{,}000$ Habitat records and $5{,}240$ ET-Plan repair records. The two bases diverge sharply on this release: task planning takes $25.00\%$ of physical records but $63.76\%$ of assistant supervision tokens, because its labels are long structured plans, while point supervision takes $25.53\%$ by physical records and $29.96\%$ by assistant tokens.}
\end{table}

\begin{table}[!htbp]
\centering
\caption{Where the two stages of the released model differ. Settings shared by both stages are listed in the note; per-stage schedules are omitted.}
\label{tab:training_config_en}
\footnotesize
\renewcommand{\arraystretch}{1.0}
\setlength{\tabcolsep}{5pt}
\fitwidth{%
\begin{tabular}{lcc}
\toprule
 & \textbf{Stage A} & \textbf{Stage B} \\
 & \makecell{Cross-Dimension\\Foundation} & \makecell{Final Joint\\Post-training} \\
\midrule
\multicolumn{3}{l}{\textit{Data}} \\
\addlinespace[2pt]
Dataset & \makecell{Open-source joint\\pre-training set} & \makecell{Final Joint\\closure set} \\
Physical training records & 462,390 & 104,960 \\
Unique identities & 294,762 & 52,480 \\
Materialization & \makecell{$1.5687\times$ on average,\\1--4 per identity} & \makecell{Exactly\\$2\times$} \\
Shards & $288$ & $128 \times 820$ \\
EQA : task planning & 90.44\% : 9.56\% (rec.) & \makecell{75.00\% : 25.00\% (rec.)\\36.24\% : 63.76\% (tok.)} \\
\midrule
\multicolumn{3}{l}{\textit{Training}} \\
\addlinespace[2pt]
Initial weights & \makecell{Pretrained\\vision-language model} & \makecell{Selected Stage A\\checkpoint} \\
LR: LLM decoder & $4.0 \times 10^{-6}$ & $3.0 \times 10^{-7}$ \\
Consumed records at selection & 358,400 & 38,400 \\
Effective epochs at selection & 0.7751 & 0.3659 \\
Wall-clock time & 19h 13m & 6h 36m \\
GPU hours & $\approx$ 615 & $\approx$ 211 \\
\bottomrule
\end{tabular}%
}
\par\smallskip
\parbox{\linewidth}{\footnotesize
\textbf{Shared by both stages.} The vision encoder is frozen; the connector, the decoder and the task-conditioned residuals are trainable, the residuals sitting on the full-attention \texttt{q\_proj} and \texttt{v\_proj} modules of layers 19, 23, 27 and 31 with rank and alpha both 16. Global batch 128 (per-device 1, accumulation 4), cosine schedule, AdamW~\cite{loshchilov2017adamw} with weight decay $0.01$, sequence length $8{,}192$, BF16, ZeRO-3~\cite{rajbhandari2019zero}, 32 GPUs. Images enter at up to $524{,}288$ px and 32 per sample, video at 1.0 fps and up to 16 frames of $262{,}144$ px. Neither release carries a regenerated token-proxy report.\\[2pt]
\textbf{Reading the table.} Wall-clock time and GPU hours describe the complete job; the consumed-record and epoch rows are computed at the selected checkpoint, which is earlier than the end of the run in both stages. Measured to that checkpoint instead, the stages take 14h 59m and 2h 32m, about $479$ and $81$ GPU hours. GPU hours are wall-clock time times 32 devices, not a billing figure. Optimizer and scheduler state are reset at the stage boundary.}
\end{table}

\paragraph{Selection guardrails and targets of the released run.} The targeted post-training run of the released model set an absolute floor of $64.32$ on the nine-benchmark question-answering mean, together with retention bounds on the remaining quantities: the task-planning mean not down by more than $0.5$ points; ET-Plan, EB-ALFRED, and EB-Habitat each not down by more than $1.0$ point; ERQA, EmbSpatial, RoboVQA, and UniEQA each not down by more than $0.7$; point, multiple-choice, and numeric parse rates at or above $99.9\%$; and zero detected overlap with the formal test sets. The benchmarks the run targeted were VSI, VABench-P, Where2Place, Cosmos, and OpenEQA. Earlier rounds set their own bounds and their own targets, which are recorded with those rounds; no single tolerance is applied across the history.

\subsection{Per-Round Ledger}

\begin{table}[!htbp]
\centering
\caption{Per-round ledger. 
Weight origin makes explicit that the history is not a single continuation chain. R3 is a diagnostic round only; see Section~\ref{sec:case_c}. The R4 scores in Table~\ref{tab:round_scores_en} were measured on a same-recipe continuation chain rather than on this round's own checkpoint. Per-round hyperparameters are omitted here; those of the two released stages are given in Table~\ref{tab:training_config_en}.
}
\label{tab:round_ledger_en}
\footnotesize
\renewcommand{\arraystretch}{1.1}
\setlength{\tabcolsep}{4pt}
\begin{tabular}{p{0.07\textwidth}p{0.245\textwidth}p{0.325\textwidth}p{0.15\textwidth}p{0.135\textwidth}}
\toprule
\textbf{Round} & \textbf{Diagnosis that triggered it} & \textbf{Data revision} & \textbf{Training change} & \textbf{Weight origin} \\
\midrule
Base & Uneven capability across the twelve embodied tasks, weakest on point grounding, video spatial reasoning, and closed-loop planning & None & None & Pretrained model \\
R1 & Validate joint planning and question-answering training and the evaluation pipeline end to end & $70{,}526$ records, $42.5$M token proxy; planning to question answering about $20\!:\!80$; box labels removed and coordinates moved to $[0,1]$ & Short pilot run, sufficient only to exercise the pipeline & Restart from pretrained \\
R2 & The pilot budget was too small to support any conclusion about capability; the nine question-answering benchmarks need explicit balancing; point format, duplication, and leakage need cleaning & $288{,}028$ records, $158.20$M token proxy; constrained sampling over nine benchmarks and $21$ fine-grained capabilities; duplicates and conflicting labels removed, $1{,}524$ test-overlapping records dropped, point data rewritten & Budget raised roughly threefold & Restart from pretrained \\
R3 & Perception improved but ALFRED and Habitat remained low; one point source held $54.30\%$ of the planning budget while grounded targets numbered $14$ records $(0.047\%)$ & $463{,}120$ records, $316.01$M token proxy; grounded ALFRED and Habitat supervision added; $65\!:\!33.4\!:\!1.2\!:\!0.4$ over question answering, general planning, Habitat, ALFRED & Trained in two consecutive segments of one release & Restart from pretrained \\
R4 & R3 reported $100$ on EB-ALFRED from a summary covering $12$ of $300$ episodes, against $27$ on recomputation over all $300$; audit confirmed official test trajectories in the legacy planning pool & $463{,}071$ records, $316.01$M token proxy; three high-risk trajectory sources removed in full ($23{,}565$ records); R3 proportions and grounded improvements retained; zero box labels, broken media, and cross-task routing errors & Same recipe as the preceding round & Restart from pretrained \\
R5 & R4 improved planning only modestly and VABench-P remained low; targeted increase needed without sacrificing perception & $153{,}600$ records, $128.95$M token; VABench $14\%$, ALFRED $17\%$, Habitat $14\%$, question answering $35\%$, general planning $20\%$ by token proxy; free-space prompt and label conflict rebuilt & Short post-training pass at a reduced learning rate & Continue from the preceding round \\
R6 & Capability peaks scattered across fragmented repairs; a single clean trunk with one release and one optimizer run was needed & $160{,}000$ unique identities materialized twice, yielding $320{,}000$ physical records; the question-answering partition is split by unique records into $45\%$ fine-grained capability, $15\%$ benchmark guarantee, and $40\%$ foundational streams; planning supervision is allocated separately, from clean ALFRED and Habitat pools & Single release, single uninterrupted optimizer run & Restart from the R4 checkpoint, deliberately bypassing R5 \\
\midrule
Stage A & Consolidate the R4 recipe under the task-conditioned architecture and the frozen contract & Open-source release derived from R4, $462{,}390$ physical records (Table~\ref{tab:stage_a_components_en}) & Large-budget foundation run & Restart from pretrained \\
Stage B & Planning recovery with a question-answering guardrail & Final Joint closure release, $104{,}960$ physical records (Table~\ref{tab:stage_b_components_en}) & Short targeted pass at a much reduced learning rate & Continue from the selected Stage~A checkpoint \\
\bottomrule
\end{tabular}
\end{table}